\documentclass[pdflatex,sn-mathphys-num]{sn-jnl}% Math and Physical Sciences Numbered Reference Style
\usepackage{graphicx}%
\usepackage{multirow}%
\usepackage{amsmath,amssymb,amsfonts}%
\usepackage{amsthm}%
\usepackage{mathrsfs}%
\usepackage[title]{appendix}%
\usepackage{xcolor}%
\usepackage{textcomp}%
\usepackage{manyfoot}%
\usepackage{booktabs}%
\usepackage{algorithm}%
\usepackage{algorithmicx}%
\usepackage{algpseudocode}%
\usepackage{listings}%
\usepackage{subcaption}
\usepackage{makecell}
\usepackage{bm}
\usepackage{tabularx}
\usepackage{placeins}
\usepackage{lineno}

\theoremstyle{thmstyleone}%
\theoremstyle{thmstyletwo}%

\theoremstyle{thmstylethree}%

\begin{document}
% \linenumbers
\title[OrthoDiffusion: A Generalizable Multi-Task Diffusion Foundation Model for Musculoskeletal MRI Interpretation]{OrthoDiffusion: A Generalizable Multi-Task Diffusion Foundation Model for Musculoskeletal MRI Interpretation}

%%=============================================================%%
%% GivenName	-> \fnm{Joergen W.}
%% Particle	-> \spfx{van der} -> surname prefix
%% FamilyName	-> \sur{Ploeg}
%% Suffix	-> \sfx{IV}
%% \author*[1,2]{\fnm{Joergen W.} \spfx{van der} \sur{Ploeg} 
%%  \sfx{IV}}\email{iauthor@gmail.com}
%%=============================================================%%
\author[1]{\fnm{Tian} \sur{Lan}}
\equalcont{These authors contributed equally to this work.}

\author[2,3,4]{\fnm{Lei} \sur{Xu}}
\equalcont{These authors contributed equally to this work.}

\author[2,3,4]{\fnm{Zimu} \sur{Yuan}}
\equalcont{These authors contributed equally to this work.}
\author[2,3,4]{\fnm{Shanggui} \sur{Liu}}
\author[2,3,4]{\fnm{Jiajun} \sur{Liu}}
\author[2,3,4]{\fnm{Jiaxin} \sur{Liu}}
\author[5,6]{\fnm{Weilai} \sur{Xiang}}
\author[5,7]{\fnm{Hongyu} \sur{Yang}}
\author*[2,3,4]{\fnm{Dong} \sur{Jiang}}
\author*[1]{\fnm{Jianxin} \sur{Yin}}
\author*[2,3,4]{\fnm{Dingyu} \sur{Wang}}

\email{wang\_dingyu@pku.edu.cn}
\email{jyin@ruc.edu.cn}
\email{bysyjiangdong@126.com}
\affil[1]{\orgdiv{Center for Applied Statistics and School of Statistics}, \orgname{Renmin University of China}, \city{Beijing}, \country{China}}

%\affil[2]{\orgdiv{Center for Applied Statistics and School of Statistics}, \orgname{Renmin University of China}, \city{Beijing}, \country{China}}

\affil[2]{\orgdiv{Department of Sports Medicine}, \orgname{Peking University Third Hospital}, \orgname{Institute of Sports Medicine of Peking University}, \city{Beijing}, \country{China}}

\affil[3]{\orgdiv{Beijing Key Laboratory of Research and Translation for Drugs and Medical Devices in Precision Diagnosis and Treatment of Sports Injuries}, \city{Beijing}, \country{China}}

\affil[4]{\orgdiv{Engineering Research Center of Sports Trauma Treatment Technology and Devices}, \orgname{Ministry of Education}, \city{Beijing}, \country{China}}

\affil[5]{\orgdiv{State Key Laboratory of Virtual Reality Technology and Systems}, \orgname{Beihang University}, \city{Beijing}, \country{China}}

\affil[6]{\orgdiv{School of Computer Science and Engineering}, \orgname{Beihang University}, \city{Beijing}, \country{China}}

\affil[7]{\orgdiv{Institute of Artificial Intelligence}, \orgname{Beihang University}, \city{Beijing}, \country{China}}

%%==================================%%
%% Sample for unstructured abstract %%
%%==================================%%

\abstract{Musculoskeletal disorders represent a significant global health burden and are a leading cause of disability worldwide. While magnetic resonance imaging (MRI) is essential for accurate diagnosis, its interpretation remains exceptionally challenging. Radiologists must identify multiple potential abnormalities within complex anatomical structures across different imaging planes, a process that requires significant expertise and is prone to variability.  To address these limitations, we developed OrthoDiffusion, a unified diffusion-based foundation model designed for multi-task musculoskeletal MRI interpretation. The framework utilizes three orientation-specific 3D diffusion models, pre-trained in a self-supervised manner on 15,948 unlabeled knee MRI scans, to learn robust anatomical features from sagittal, coronal, and axial views. These view-specific representations are integrated to support diverse clinical tasks, including anatomical segmentation and multi-label diagnosis. Our evaluation demonstrates that OrthoDiffusion achieves excellent performance in the segmentation of 11 knee structures and the detection of 8 knee abnormalities. The model exhibited remarkable robustness across different clinical centers and MRI field strengths, consistently outperforming traditional supervised models. Notably, in settings where labeled data was scarce, OrthoDiffusion maintained high diagnostic precision using only 10\% of training labels. Furthermore, the anatomical representations learned from knee imaging proved highly transferable to other joints, achieving strong diagnostic performance across 11 diseases of the ankle and shoulder. These results highlight the model's capacity for cross-anatomical generalization without the need for extensive joint-specific retraining. These findings suggest that diffusion-based foundation models can serve as a unified platform for multi-disease diagnosis and anatomical segmentation, potentially improving the efficiency and accuracy of musculoskeletal MRI interpretation in real-world clinical workflows.}

%%================================%%
%% Sample for structured abstract %%
%%================================%%

\keywords{Musculoskeletal MRI, Diffusion models, Self-supervised representation learning, Multi-plane fusion, Multi-label diagnosis, Anatomical segmentation, Cross-anatomical generalization, Label-efficient learning, Unified imaging framework}

%%\pacs[JEL Classification]{D8, H51}

%%\pacs[MSC Classification]{35A01, 65L10, 65L12, 65L20, 65L70}

\maketitle

\section{Introduction}\label{sec:Introduction}
Musculoskeletal disorders represent a leading cause of disability worldwide, affecting approximately 1.71 billion of patients~\cite{Musahl2019ACL, Ouyang2025GBD, Poulsen2019KOA, WHO2022_MSK}. Magnetic Resonance Imaging (MRI) serves as the gold-standard for non-invasive evaluation, providing high-fidelity soft-tissue contrast across multiple anatomical sites~\cite{Kijowski2014Imaging,Nacey2017MRI}. However, the interpretation of musculoskeletal MRI remains clinically challenging and time-consuming due to several inherent complexities, including the frequent coexistence of multiple abnormalities within a single examination, intricate three-dimensional anatomy, and the need to integrate complementary diagnostic information from multiple imaging planes. Although deep-learning systems have shown promise in detecting isolated conditions such as anterior cruciate ligament injuries~\cite{Wang2024ACLDeepLearning, Tran2022ACLDetection}, meniscal lesions~\cite{Botnari2024MeniscusReview}, and rotator cuff tears~\cite{Lin2023RotatorCuffDL}, among others, most existing approaches are narrowly designed for single diagnostic tasks at specific anatomical joints. Trained on limited data, these models often struggle to generalize across different clinical centers, MRI scanner protocols, and diverse patient populations, limiting their practical utility in real-world clinical workflows~\cite{Sun2025KneeInjuryDetection}.

Recent advances in foundation models, particularly through self-supervised learning~\cite{devlin-etal-2019-bert, Radford2018ImprovingLU, Raffel2019ExploringTL, He2021MaskedAA, Chen2020GenerativePF}, offer a promising pathway toward more generalizable medical image analysis~\cite{yan2025panderm, Jiang2025SilentInfarction, Christensen2024EchoVLM, Zhou2023RetinalFoundation, deltadahl2025deep}. Among these, diffusion models have emerged as powerful generative frameworks that learn rich data distributions by iteratively denoising corrupted inputs~\cite{Ho2020DenoisingDP, Song2020ScoreBasedGM, Wang2025SelfImproving, Bluethgen2024ChestXrayGen}. Beyond image synthesis, the intermediate representations learned during this denoising process capture robust, multi-scale features~\cite{ddae2023, Baranchuk2021LabelEfficientSS}. Our previous study demonstrated that diffusion pre-training as a unified approach to simultaneously acquire generation ability and deep visual understandings, which potentially leads to the development of unified vision foundation models for diverse downstream clinical tasks~\cite{ddae2023}. Compared with discriminative learning approaches that optimize task-specific objectives~\cite{Li2022IdentificationAD,Roblot2019ArtificialIT,Fritz2020MeniscusTears,Li2022MeniscusSegmentation,Rizk2021MeniscalLD,Singh20203DDL,Dosovitskiy2020AnII}, the multi-step denoising process of diffusion models inherently captures hierarchical features, providing a continuous spectrum of feature abstraction that allows a single model to adapt flexibly to different diagnostic tasks without architectural modifications~\cite{Rombach2021HighResolutionIS, Luo2022UnderstandingDM}. This generative denoising objective also promotes resilience to the types of noise and variability commonly encountered across clinical centers and MRI scanners.

\begin{figure}[tp]
    \centering
    \includegraphics[width=0.78\linewidth]{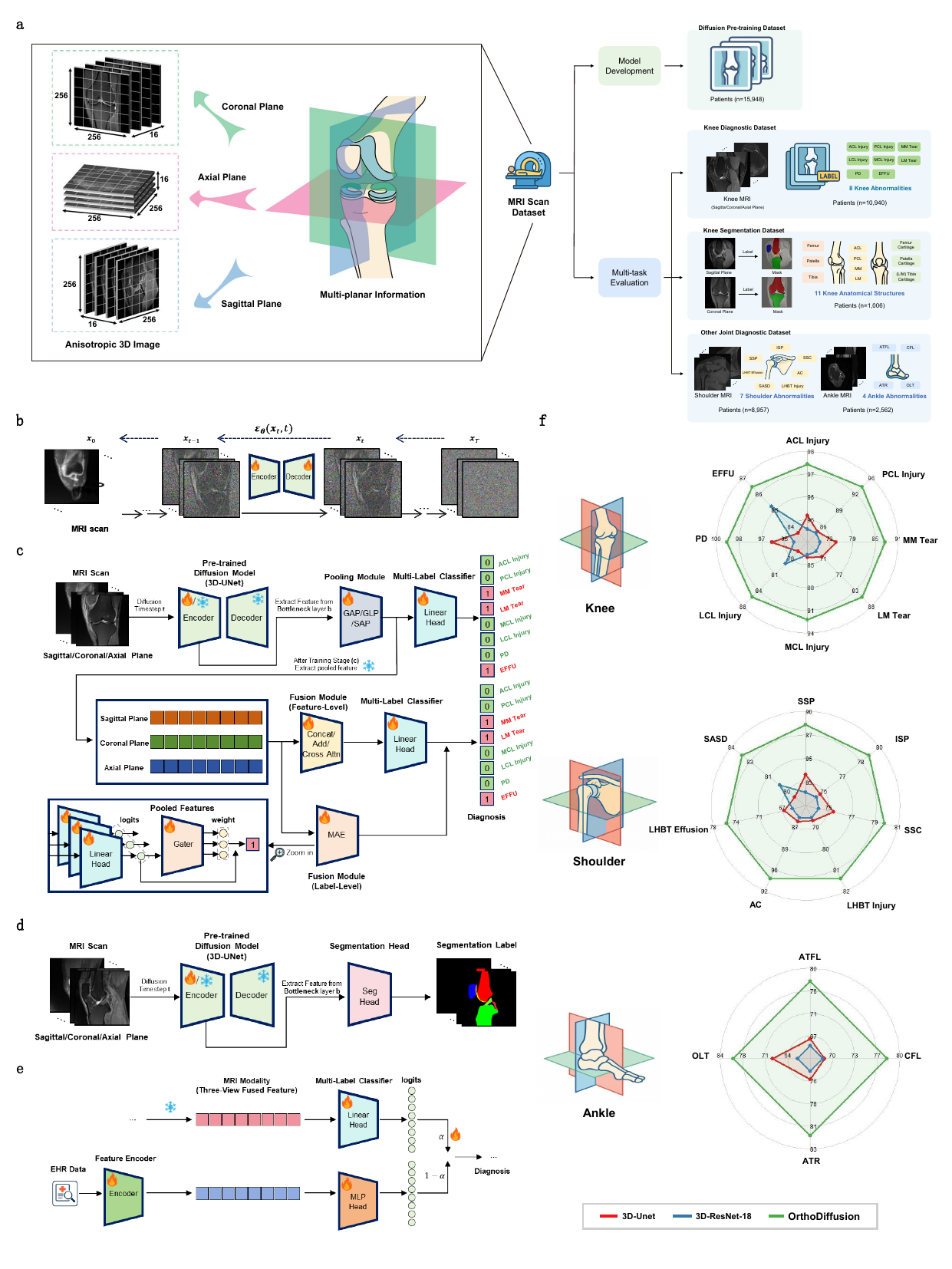}
    \caption{\textbf{Overview of the OrthoDiffusion framework.} 
    \textbf{a}, Multi-planar musculoskeletal MRI acquisition, including sagittal, coronal, and axial views, with anisotropic resolution across slices. Dataset construction and task composition across joints are illustrated.
    \textbf{b}, Unconditional 3D diffusion pretraining using a 3D U-Net noise predictor on large-scale knee MRI data.
    \textbf{c}, Feature extraction from intermediate diffusion representations at selected timesteps and bottleneck blocks, followed by pooling and multi-label classification. Feature-level and label-level fusion strategies are applied to integrate sagittal, coronal, and axial representations.
    \textbf{d}, Anatomical segmentation pipeline using diffusion representations coupled with a lightweight segmentation head.
    \textbf{e}, Multimodal fusion strategy integrating MRI diffusion representations with structured electronic health record (EHR) data for diagnosis.
    \textbf{f}, Diagnostic performance (AUROC) across 19 musculoskeletal abnormalities involving the knee, ankle, and shoulder joints, compared with baseline CNN models.}
    \label{fig:overview}
\end{figure}
To translate these advances into clinical practice, we propose OrthoDiffusion, a diffusion-based representation learning framework for scalable musculoskeletal MRI interpretation. Mirroring the multi-plane analysis routinely performed by radiologists, OrthoDiffusion employs three independent, orientation-specific 3D diffusion branches, each pretrained on large-scale unlabeled MRI data, to learn anatomy-aware features from sagittal, coronal, and axial views. These view-specific representations are subsequently integrated through an adaptive fusion strategy to support heterogeneous downstream tasks, including multi-label diagnosis and anatomical segmentation across different joints. We constructed one of the largest multi-joint musculoskeletal MRI datasets, comprising over 1.44 million images and 91,959 sequences, from 30,653 MRI examinations of the knee, ankle, and shoulder (Fig.~\ref{fig:overview}(a)). Pretrained on 15,948 knee MRI examinations, OrthoDiffusion achieves robust, label-efficient performance on 11 knee anatomic structures segmentation and 8 knee diseases diagnosis across multiple clinical centers and MRI field strengths. Besides, we further demonstrated that OrthoDiffusion exhibits strong cross-anatomy transferability on other joints, with 7 diagnostic tasks on shoulder and 4 tasks in ankle (Fig.~\ref{fig:overview}(f)). Crucially, this is achieved through a single unified model with one fine-tuning, supporting concurrent multi-label diagnosis and anatomical segmentation without requiring separate task-specific fine-tuning. This unified, diffusion-based paradigm offers a scalable and clinically coherent framework for advancing musculoskeletal MRI interpretation toward broader real-world adoption.

\section{Results}\label{sec:results}

\subsection{Overview of the study}
The overall workflow of the study is illustrated in Fig.~\ref{fig:overview}(b)-(e). The proposed framework was developed and validated using a large-scale, multi-center MRI cohort consisting of over 30,000 patients from nine clinical institutions. The dataset covers the knee, ankle, and shoulder joints, featuring annotations for 11 anatomical structures and 19 clinically relevant musculoskeletal abnormalities. Detailed task definitions and abbreviation conventions are provided in Table~\ref{tab:msk_tasks}.

\begin{table}[htbp]
\centering
\caption{Unified definitions of anatomical structures and abnormalities for musculoskeletal MRI tasks. Abbreviations are defined separately for segmentation and classification.}
\label{tab:msk_tasks}
\begin{tabular}{l l p{7.5cm}}

\toprule
\textbf{Task} & \textbf{Joint} & \textbf{Structure / Abnormality (Abbreviation)} \\
\midrule
\multirow{11}{*}{Segmentation}
& \multirow{11}{*}{Knee}
& Femur  \\
& & Tibia \\
& & Patella \\
& & Femoral cartilage (FC) \\
& & Medial tibial cartilage (MTC) \\
& & Lateral tibial cartilage (LTC) \\
& & Medial meniscus (MM) \\
& & Lateral meniscus (LM) \\
& & Anterior cruciate ligament (ACL) \\
& & Posterior cruciate ligament (PCL) \\
& & Patellar cartilage (PC) \\
\midrule

\multirow{19}{*}{Classification}
& \multirow{8}{*}{Knee}
& Anterior cruciate ligament (ACL) injury  \\
& & Posterior cruciate ligament (PCL) injury \\
& & Medial meniscus (MM) tear \\
& & Lateral meniscus (LM) tear \\
& & Medial collateral ligament (MCL) injury \\
& & Lateral collateral ligament (LCL) injury \\
& & Patellar dislocation (PD) \\
& & Joint effusion (EFFU) \\
\cmidrule(lr){2-3}

& \multirow{4}{*}{Ankle}
& Anterior talofibular ligament (ATFL) injury \\
& & Calcaneofibular ligament (CFL) injury \\
& & Achilles tendon rupture (ATR) \\
& & Osteochondral lesion of the talus (OLT) \\
\cmidrule(lr){2-3}

& \multirow{7}{*}{Shoulder}
& Supraspinatus tendon (SSP) tear   \\
& & Infraspinatus tendon (ISP) tear \\
& & Subscapularis tendon (SSC) tear \\
& & Long head of the biceps tendon (LHBT) injury \\
& & Adhesive capsulitis (AC) \\
& & Long head of the biceps tendon (LHBT) sheath effusion \\
& & Subacromial-subdeltoid (SASD) bursal effusion  \\
\bottomrule
\end{tabular}
\end{table}

\begin{table}[htbp]
    \centering
    \setlength{\tabcolsep}{1pt}
    \caption{Segmentation performance of OrthoDiffusion under less than 30\% labeled data (Dice Similarity Coefficient, \%) for 11 \textbf{knee anatomical structures} on sagittal and coronal MRI scans. Entries marked as ``--'' indicate anatomical structures that are not visible in the coronal view (e.g., Patella and PC). Boldface indicates the best performance under the same setting in each column.}
    \label{tab:seg_dice}
    \begin{tabular}{llccccccccccc}
        \toprule
        \textbf{Orientation} & \textbf{Methods}  & \textbf{Femur} & \textbf{Tibia} & \textbf{Patella} & \textbf{FC} & \textbf{MTC} & \textbf{LTC} & \textbf{PC} & \textbf{MM} & \textbf{LM} & \textbf{ACL} & \textbf{PCL} \\
        \midrule
        \multirow{3}{*}{Sagittal} 
        & 3D-Unet & 91.91 & 90.67 & 82.25 & \textbf{64.45} & 50.27 & 45.54 & \textbf{65.59} & 50.71 & 53.13 & 46.49 & 51.40 \\
        & UNETR & 83.32 & 73.90 & 57.05 & 44.84 & 40.99 & 26.85 & 48.54 & 30.12 & 21.14 & 30.74 & 15.30  \\
        & \textbf{Ours (FT, FT-Opt-Sag)} & \textbf{93.02} & \textbf{92.54} & \textbf{84.58} & 58.50 & \textbf{63.76} & \textbf{63.07} & 64.13 & \textbf{60.69} & \textbf{56.03} & \textbf{53.67} & \textbf{58.34} \\
        \midrule
        \multirow{3}{*}{Coronal} 
        & 3D-Unet & 90.87 & 90.54 & -- & 61.12 & 68.51 & 64.63 & -- & 63.49 & 58.40 & 45.72 & 42.19 \\
        & UNETR & 87.30 & 84.37 & -- & 58.87 & 57.61 & 49.50 & -- & 46.87 & 43.86 & 38.96 & 46.66  \\
        & \textbf{Ours (FT, FT-Opt-Cor)} & \textbf{92.91} & \textbf{92.82} & -- & \textbf{64.21} & \textbf{69.73} & \textbf{69.63} & -- & \textbf{68.37} & \textbf{65.33} & \textbf{64.13} & \textbf{60.06} \\
        \bottomrule
    \end{tabular}
\end{table}

We utilized 3D diffusion-based architecture for representation learning, incorporating orientation-specific pretraining across sagittal, coronal, and axial planes. Feature representations extracted from selected diffusion timesteps and network blocks were employed for downstream tasks, including anatomical segmentation and disease classification. 

The results demonstrated that OrthoDiffusion consistently outperformed supervised baselines, particularly in limited-label situation, while maintaining robustness across different joints and heterogeneous imaging conditions.

\subsection{Performance on Anatomical Segmentation}

The model’s capability of the learned diffusion representations used for dense prediction tasks was evaluated on knee MRI scans. As illustrated in Fig.~\ref{fig:overview}(d), a lightweight segmentation head was trained on top of features extracted from the pretrained OrthoDiffusion framework to perform segmentation tasks of 11 anatomical structures. 

\begin{figure}[t]
    \centering
    \includegraphics[width=\linewidth]{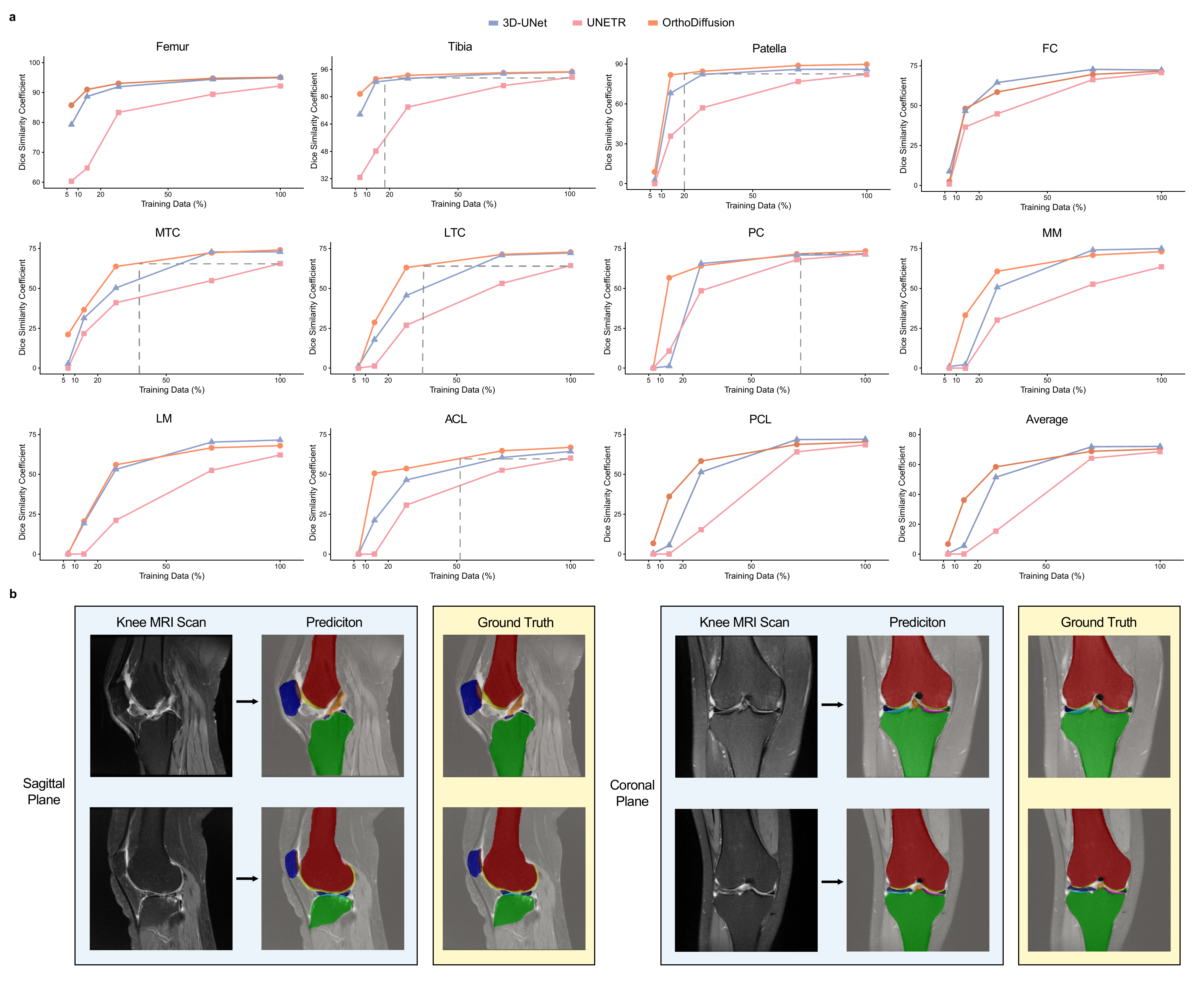}
    \caption{\textbf{Qualitative and quantitative evaluation of OrthoDiffusion for knee anatomical segmentation.} \textbf{a}, Label efficiency of OrthoDiffusion for knee anatomical segmentation \textbf{on the Sagittal plane}, illustrated by representative Dice Similarity Coefficient curves for selected anatomical structures under progressively reduced supervision. \textbf{b}, Qualitative multi-planar segmentation results. Representative sagittal and coronal knee MRI slices showing the original input images, corresponding model-predicted anatomical segmentations (e.g., femur, tibia, cartilage, etc.), and manual ground-truth annotations, illustrating accurate delineation of key joint structures across imaging planes.}
    \label{fig:seg_vis}
\end{figure}

The results indicate that the proposed framework, with minimal fine-tuning, consistently outperformed strong CNN-based and Transformer-based models~\cite{iek20163DUL,hatamizadeh2022unetr} trained from scratch under identical experimental settings in the limited-label regime (Table~\ref{tab:seg_dice} and Fig.~\ref{fig:seg_vis}(a)). Qualitative visualizations in Fig.~\ref{fig:seg_vis}(b) further demonstrate the anatomical fidelity of the segmentation masks. These findings suggest that the diffusion-based representations learned by OrthoDiffusion effectively capture anatomically meaningful structural and textural information. Further implementation details are provided in the Supplementary Material.
% We evaluated the performance of diffusion representations learned by OrthoDiff on dense prediction tasks. As illustrated in Fig.~\ref{fig:overview} (d), a lightweight segmentation head was trained on top of features extracted from the pretrained OrthoDiff framework to perform multi-class segmentation tasks of 11 anatomical structures in knee MRI sequences. 

% With minimal fine-tuning of the pretrained diffusion backbone, OrthoDiff consistently outperformed classical CNN-based segmentation models~\cite{iek20163DUL,hatamizadeh2022unetr} trained from scratch under identical experimental settings in the limited-label regime, as shown in Fig.~\ref{fig:label_efficiency_cross_anatomy} (b). Qualitative visualizations in Fig.~\ref{fig:seg_vis} further demonstrate the anatomical fidelity of the segmentation masks. 

% These findings indicate that the diffusion-based representations learned by OrthoDiff capture  anatomically  meaningful  structural  and  textural information. Further implementation details are provided in the Supplementary Material.

\subsection{Diagnostic Accuracy and Robustness in Knee MRI}

% We adapted two baseline architectures—3D-UNet~\cite{iek20163DUL} and 3D-ResNet-18~\cite{Hara2017CanS3}-to our diagnostic task to benchmark performance. These architectures were selected as they represent commonly used 3D CNN backbones for medical image analysis. 
%The diagnostic task was formulated as a multi-label classification problem, with the area under the receiver operating characteristic curve (AUC-ROC) used as the primary evaluation metric. Additional evaluation metrics, including mean average precision (mAP), per-class precision (CP), per-class recall (CR), per-class F1 score (CF1), overall precision (OP), overall recall (OR), and overall F1 score (OF1), are provided in the Supplementary Materials.

As summarized in Table~\ref{tab:knee_multilabel_aucroc_merged} and Fig.~\ref{fig:overview}(f), the optimal configuration, which selected specific diffusion timesteps and bottleneck blocks based on validation performance, enabled OrthoDiffusion to achieve a macro-average AUROC of 0.908 across eight knee abnormality categories, outperforming baseline approaches~\cite{iek20163DUL,Hara2017CanS3}.

\begin{table}[htbp]
    \centering
    \setlength{\tabcolsep}{1pt}
    \caption{Comparison of per-class AUROC (\%) for the eight-label \textbf{knee injury prediction} task using different modalities (MRI-only, EHR-only, and MRI+EHR) on the Center A+B+C+D+E+F+G test set. Boldface indicates the best performance under the same setting for each column.}
    \label{tab:knee_multilabel_aucroc_merged}
    \begin{tabular}{l l l ccccccccc}
        \toprule
        \textbf{Dataset} & \textbf{Modality} & \textbf{Methods} 
        & \textbf{ACL} & \textbf{PCL} & \textbf{MM} & \textbf{LM} & \textbf{MCL} & \textbf{LCL} & \textbf{PD} & \textbf{EFFU} & \textbf{Macro-AUC} \\
        \midrule

        \multirow{9}{*}{\shortstack{Center A+B+\\C+D+E+F+G}} 

        % -------- MRI only --------
        & \multirow{4}{*}{MRI-only}
        & 3D-Unet 
        & 94.35 & 84.76 & 70.91 & 68.76 & 83.44 & 79.13 & 94.67 & 85.03 & 82.63 \\
        & & 3D-ResNet-18 & 94.91 & 85.00 & 74.87 & 70.74 & 83.71 & 76.63 & 96.17 & 83.21 & 83.15 \\
        & & \textbf{Ours (LP, LP-Opt)} & 95.15 & 91.26 & 81.37 & 76.23 & 89.08 & 82.58 & 98.85 & 86.34 & 87.61 \\
        % & \textbf{Ours (Fine-tuning, LP-Opt)} & \textbf{98.16} & \textbf{94.47} & 86.56 & 84.52 & 92.72 & \textbf{85.98} & 99.61 & \textbf{86.43} & 91.18 \\
        & & \textbf{Ours (FT, FT-Opt)} & \textbf{97.04} & \textbf{93.73} & \textbf{87.24} & \textbf{85.02} & \textbf{91.91} & \textbf{85.77} & \textbf{99.26} & \textbf{86.35} & \textbf{90.79} \\

        \cmidrule(l){2-12}

        % -------- EHR only --------
        & EHR-only
        & EHR-only
        & 52.37 & 53.22 & 48.77 & 50.19 & 54.24 & 62.02 & 57.88 & 55.99 & 54.33 \\

        \cmidrule(l){2-12}

        % -------- Multi-modal --------
        & \multirow{2}{*}{EHR + MRI}
        & \textbf{Ours (LP, LP-Opt)}
        & 95.24 & 91.23 & 81.40 & 76.38 & 89.24 & 84.29 & 98.84 & 86.84 & 87.93 \\

        & & \textbf{Ours (FT, FT-Opt)}
        & \textbf{97.02} & \textbf{93.80} & \textbf{87.21} & \textbf{85.09} & \textbf{92.01} & \textbf{86.56} & \textbf{99.27} & \textbf{86.95} & \textbf{90.99} \\

        \bottomrule
    \end{tabular}
\end{table}

\begin{table}[htbp]
    \centering
    \setlength{\tabcolsep}{2pt}
    \caption{Comparisons of per-disease AUROC (\%) for the eight-label \textbf{knee injury prediction} task on the Center A+B+C+D+E+F+G test set with different magnetic field strengths. Boldface indicates the best result under the same setting in each column.}
    \label{tab:knee_multilabel_baseline_aucroc_field_strengths_1_3}
    \begin{tabular}{l l ccccccccc}
        \toprule
        \textbf{Field Strength} & \textbf{Methods} & \textbf{ACL} & \textbf{PCL} & \textbf{MM} & \textbf{LM} & \textbf{MCL} & \textbf{LCL} & \textbf{PD} & \textbf{EFFU} & \textbf{Macro-AUC} \\
        \midrule
        \multirow{3}{*}{1.5T} 
        & 3D-Unet & 94.06 & 84.06 & 70.60 & 71.42 & 84.72 & 82.76 & 91.44 & 88.13 & 83.40  \\
        & 3D-ResNet-18 & 93.76 & 85.50 & 75.13 & 71.75 & 86.35 & 82.44 & 94.84 & 85.64 & 84.43  \\
        & \textbf{Ours (FT, FT-Opt)} & \textbf{96.87} & \textbf{94.99} & \textbf{88.49} & \textbf{83.50} & \textbf{93.66} & \textbf{89.88} & \textbf{98.78} & \textbf{88.96} & \textbf{91.89}\\
        \midrule
        \multirow{3}{*}{3T}
        & 3D-Unet 
        & 94.57 & 85.31 & 71.06 & 67.95 & 83.18 & 77.24 & 95.78 & 84.28 & 82.42 \\
        
        & 3D-ResNet-18 
        &  95.76 & 85.91 & 74.80 & 70.51 & 81.14 & 75.24 & 97.03 & 82.98 & 82.92 \\
        
        & \textbf{Ours (FT, FT-Opt)} 
        & \textbf{97.29} & \textbf{93.99} & \textbf{87.07} & \textbf{85.36} & \textbf{91.11} & \textbf{84.54} & \textbf{99.40} & \textbf{85.51} & \textbf{90.53} \\
        \bottomrule
    \end{tabular}
\end{table}
% To further assess model robustness, we evaluated the trained model on the external multi-center dataset. The model trained on the internal dataset was applied directly to the external cohort under matched PD-FS sequence configurations. As shown in Table~\ref{tab:knee_multilabel_six_baseline_aucroc}, the linear probing configuration showed the strongest robustness, with performance decreasing by approximately 6\%. In contrast, both baseline models experienced drops exceeding 11\%, underscoring the superior robustness of C-UNIDIFF.

% Notably, the linear probing configuration exhibited stronger robustness on the external dataset than the fine-tuned variants. This observation is consistent with prior findings that fine-tuning may overfit to the source-domain distribution~\cite{Kumar2022FineTuningCD}, thereby reducing cross-domain robustness. The divergence in performance further indicates a distributional shift between the internal and external datasets, likely attributable to differences in scanning parameters and patient demographics. The diffusion-based representation provides a principled and effective solution to mitigate this domain mismatch, enabling more stable diagnostic performance across diverse clinical environments.

\begin{figure}[tp]
    \centering
    \includegraphics[width=\linewidth]{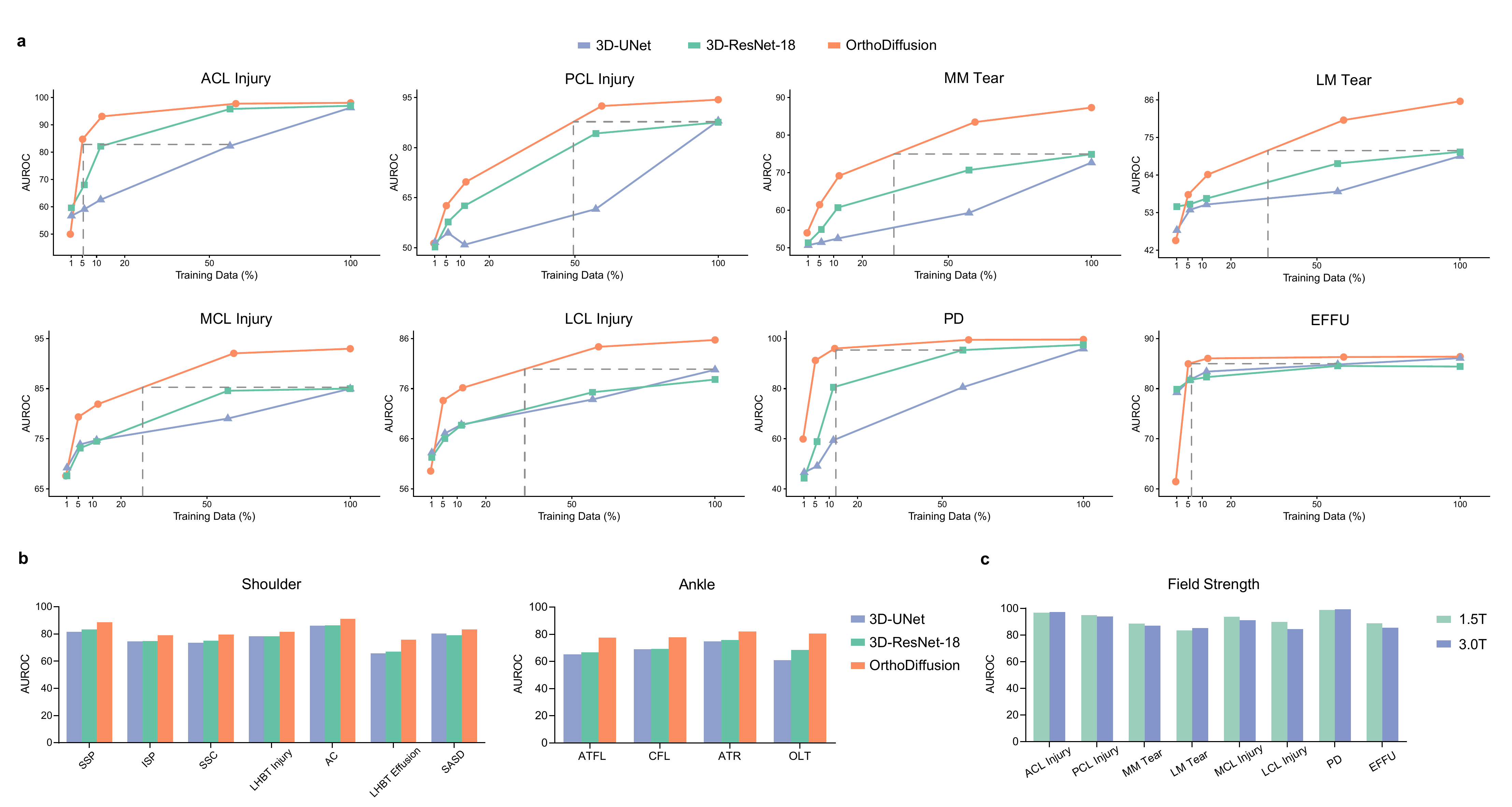}
    \caption{\textbf{Label efficiency, cross-anatomy transferability, and robustness of OrthoDiffusion.} 
    \textbf{a}, Diagnostic performance (AUROC) of OrthoDiffusion on eight knee abnormalities as a function of the proportion of labeled training data, demonstrating label-efficient learning under limited supervision.
    \textbf{b}, Cross-anatomy generalization performance of OrthoDiffusion on ankle and shoulder abnormality diagnosis, transferred from a diffusion backbone pretrained exclusively on knee MRI.
    \textbf{c}, Robustness of OrthoDiffusion to variations in MRI magnetic field strength (1.5T and 3.0 T), evaluated on knee MRI acquired at different field strengths.
    }
    
    \label{fig:label_efficiency_cross_anatomy}
\end{figure}

Additionally, OrthoDiffusion demonstrated robustness to domain shifts induced by variations in magnetic field strength (Table~\ref{tab:knee_multilabel_baseline_aucroc_field_strengths_1_3}, Supplementary Table~\ref{tab:knee_multilabel_baseline_ap_field_strengths}, and Fig.~\ref{fig:label_efficiency_cross_anatomy}(c)). 
% In addition, OrthoDiff demonstrated robustness to Table~\ref{tab:knee_multilabel_baseline_aucroc_field_strengths_1_3} and Fig.~\ref{fig:label_efficiency_cross_anatomy} (d) report OrthoDiff performance across MRI datasets acquired at different magnetic field strengths, highlighting its robustness to field strength–induced domain shifts.

An exploratory multi-modal analysis integrating structured Electronic Health Records (EHR) with MRI representations suggested that multi-modal data provides complementary diagnostic cues, yielding modest performance gains (Table~\ref{tab:knee_multilabel_aucroc_merged} and Supplementary Table~\ref{tab:knee_multilabel_ap_merged}).
% To assess the contribution of complementary clinical information, we conducted an exploratory multi-modal analysis integrating structured Electronic Health Records (EHR) data with MRI-based diffusion representations, as summarized in Table~\ref{tab:knee_multilabel_aucroc_merged}. The results suggest that incorporating structured EHR information provides complementary diagnostic cues, although the overall performance gains are modest.

Furthermore, OrthoDiffusion exhibited strong label efficiency. As shown in Fig.~\ref{fig:label_efficiency_cross_anatomy}(a), it retained high diagnostic precision even when trained with only 10\% of the labeled data, showing significantly less performance degradation compared to CNN baselines. This label efficiency demonstrates that the diffusion model captures high-quality, anatomically meaningful representations that can be leveraged effectively with minimal downstream supervision. 

% We also assessed the ability of OrthoDiff to perform accurate diagnosis under progressively reduced supervision. As shown in Fig.~\ref{fig:label_efficiency_cross_anatomy} (a), OrthoDiff retained strong diagnostic performance even with substantially fewer labels. Notably, when trained with only 10\% of the labeled data, the performance degradation remained modest compared with CNN baselines. This label efficiency demonstrates that the diffusion model captures high-quality, anatomically meaningful representations that can be leveraged effectively with minimal downstream supervision—an essential property for foundation models in medical imaging, where labeled data are costly and difficult to obtain.

\subsection{Cross-Anatomy Generalization of OrthoDiffusion}

% To further assess the scalability and transferability of OrthoDiff as a unified framework, we evaluated its cross-anatomy generalization across distinct downstream tasks.

% We first assessed the ability of C-UNIDIFF to perform accurate diagnosis under progressively reduced supervision (details in Supplementary Materials). As shown in Fig.~\ref{fig:label_efficiency}, C-UNIDIFF retained strong diagnostic performance even with substantially fewer labels. Notably, when trained with only 10\% of the labeled data, the performance degradation remained modest compared with CNN baselines. This label efficiency demonstrates that the diffusion model captures high-quality, anatomically meaningful representations that can be leveraged effectively with minimal downstream supervision—an essential property for foundation models in medical imaging, where labeled data are costly and difficult to obtain.

To evaluate scalability and transferability, the diffusion backbone of OrthoDiffusion, which pretrained exclusively on \textbf{knee} MRI, was applied to \textbf{ankle} and \textbf{shoulder} classification tasks. Despite substantial morphological differences across joints, OrthoDiffusion achieved competitive diagnostic performance on ankle and shoulder abnormalities with minimal fine-tuning (Fig.~\ref{fig:overview}(f), Fig.~\ref{fig:label_efficiency_cross_anatomy}(b), and Supplementary Table~\ref{tab:ankle_multicenter_ap}-\ref{tab:shoulder_multicenter_auc}). OrthoDiffusion consistently outperformed 3D CNN baselines trained from scratch for each specific anatomy. These findings highlight that the backbone learns anatomical and transferable pathological representations that are not confined to a single joint. Moreover, when target joints share higher anatomical and imaging-level similarity, such as the knee and ankle, the diffusion backbone transfers richer medical priorities, resulting in more pronounced performance improvements.

\subsection{Interpretable Multi-Plane Fusion Strategies}
Our study systematically evaluated fusion strategies for integrating sagittal, coronal, and axial MRI representations. While simple concatenation yielded the highest macro-average AUROC (Supplementary Table~\ref{tab:linearprobing-fusion-aucroc} and~\ref{tab:linearprobing-fusion-multilabel}), it lacks transparency and interpretability regarding the contribution of each anatomical plane. To address this, we proposed a Multi-plane Adaptive Expert (MPAE) Fusion framework. 

MPAE operates by treating the sagittal, coronal, and axial branches as three independent experts, each producing disease-specific predictions for downstream fusion. For each diagnostic label, these outputs are passed into an adaptive gating mechanism that computes patient-specific, orientation-wise fusion weights. After normalization, these weights determine the relative contribution of each anatomical plane to the final diagnostic output. 

\begin{figure}[t]
    \centering
    \includegraphics[width=\linewidth]{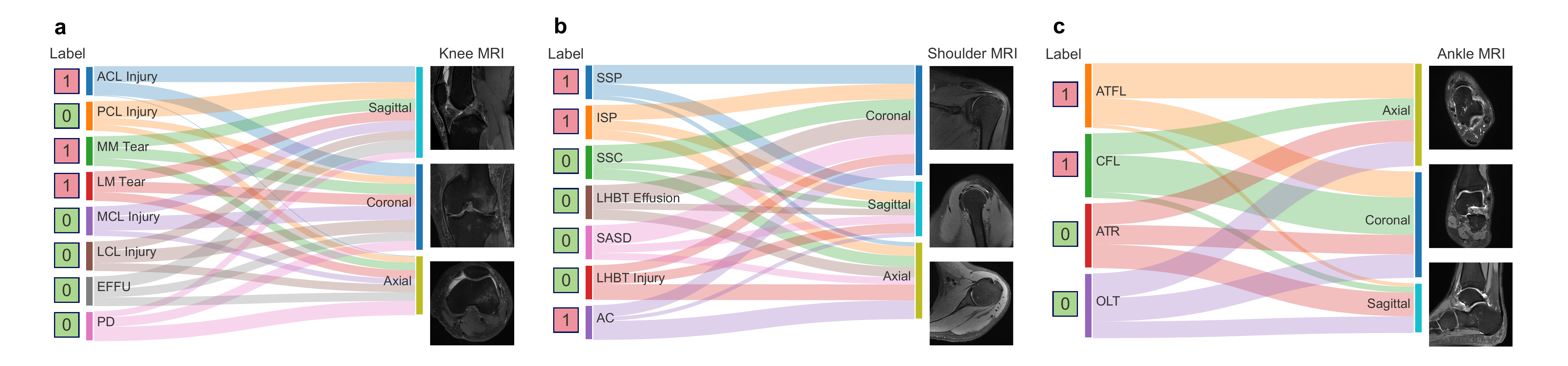}
    \caption{\textbf{Orientation-specific fusion expert-weights predicted by MPAE reflect clinical imaging practice.} Sankey diagrams showing label-specific fusion weights assigned by the MPAE module across axial, sagittal, and coronal planes for representative (a) knee, (b) shoulder, and (c) ankle abnormalities in a single patient. The relative thickness of each flow indicates the contribution of each imaging orientation to the final prediction, revealing plane preferences that align with established clinical reading protocols for different orientations and pathologies. Representative MRI slices from each orientation are shown for reference.}
    \label{fig:mae_weights}
\end{figure}

Although simple concatenation was used as the default for maximum numerical performance, MPAE provides clinically relevant interpretability. Visualization of the learned fusion weights via Sankey diagrams (Figure~\ref{fig:mae_weights}) reveals that the model prioritizes specific planes for different disorders, such as relying on sagittal views for cruciate ligament injuries and coronal views for collateral ligament injuries, aligning with real-world clinical practice.

\section{Discussion}\label{Discussion}

This study introduces OrthoDiffusion, a unified diffusion-based representation learning framework that addresses enduring challenges in musculoskeletal MRI interpretation, such as the frequent co-occurrence of multiple abnormalities, complex three-dimensional anatomy, and the need to integrate complementary information from multiple imaging planes. By employing orientation-specific 3D diffusion backbones pre-trained on a large-scale cohort of 15,948 knee MRI examinations, OrthoDiffusion demonstrates robust performance across diverse joints and heterogeneous clinical tasks—ranging from anatomical segmentation to multi-label diagnosis—using a single unified model. These results underscore the potential of diffusion models to serve not only as generative tools but also as powerful foundational backbones for label-efficient and generalizable medical image analysis.

The effectiveness of OrthoDiffusion stems from its self-supervised pre-training paradigm, which leverages extensive unlabeled MRI data to learn transferable anatomical priors. Unlike masked autoencoding~\cite{He2021MaskedAA, xie2022simmimsimpleframeworkmasked}, which excels at learning spatial context through reconstruction, or contrastive learning~\cite{chen2020simple, chen2020bigselfsupervisedmodelsstrong, grill2020bootstraplatentnewapproach, caron2021emergingpropertiesselfsupervisedvision}, which builds invariant representations by discriminating between instances, diffusion models~\cite{Ho2020DenoisingDP, Song2020ScoreBasedGM} adopt a generative approach. Their multi-step denoising process inherently captures hierarchical features across continuous noise levels, yielding a flexible spectrum of abstractions that can be dynamically adapted to various downstream tasks without modifying the network architecture. Although diffusion pre-training is computationally demanding and requires careful tuning of noise schedules, its capacity to model complex three-dimensional data distributions and produce noise-robust representations offers a compelling advantage. The denoising objective itself promotes inherent robustness to imaging variations arising from different MRI scanners, acquisition protocols, and magnetic field strengths, thereby enhancing generalization across clinical sites. Moreover, the diffusion timestep provides an additional dimension of representational flexibility~\cite{ddae2023, Baranchuk2021LabelEfficientSS, Li2025UnderstandingRD, tang2023emergentcorrespondenceimagediffusion}. By modulating the noise level at which features are extracted, OrthoDiffusion enables different downstream tasks to access representations at varying levels of abstraction, offering a principled mechanism for selecting task-adaptive features without altering the underlying model (Supplementary Table~\ref{tab:timestep_block_selection} and Extended Data Fig.~\ref{fig:aucroc_comparison}). This property makes diffusion modeling an promising foundation for a unified framework capable of supporting concurrent multi-label diagnosis and anatomical segmentation.

Effective self-supervised pre-training typically requires large and diverse datasets, which have been relatively scarce in musculoskeletal MRI compared with other medical imaging domains. While public repositories and foundation models have accelerated progress in specialties such as ophthalmology~\cite{Zhou2023RetinalFoundation, Sun2025DataEfficientMedical}, pathology~\cite{Ding2025MultimodalPathologyFoundation, Ma2025GeneralizablePathologyFoundation, Xiang2025VisionLanguageOncology}, and dermatology~\cite{Yan2025MultimodalDermatologyFoundation}, musculoskeletal imaging has lagged due to challenges in data accessibility and annotation complexity. To bridge this gap, we curated one of the largest multi-joint musculoskeletal MRI cohorts to date, comprising 1.44 million images and 91,959 sequences derived from 30,653 examinations of the knee, ankle, and shoulder, with comprehensive annotations spanning 11 anatomical structures and 19 abnormalities. This dataset not only supports robust pre-training and finetuning but also enables systematic evaluation of cross-anatomy generalization.

In terms of model architecture, OrthoDiffusion is tailored to musculoskeletal MRI through three independent diffusion backbones dedicated to sagittal, coronal, and axial views, each trained to capture orientation-specific anatomical context. Through adaptive fusion, the model integrates complementary planar information in a manner that aligns with radiological reading practices. Supported by this substantial dataset, OrthoDiffusion demonstrates high label efficiency, outperforming strong baselines even when trained on only 10\% of labeled data. The framework also exhibits multi-task adaptability and cross-anatomical transferability: a backbone pre-trained on knee MRI substantially improves diagnostic performance on ankle and shoulder examinations with minimal fine-tuning. These outcomes highlight the model’s ability to learn musculoskeletal representations that are both anatomically meaningful and pathologically informative, rather than relying on task-specific heuristics, thereby supporting robust transfer across tasks and anatomical regions within a unified framework. Consequently, the pre-trained backbone can remain frozen while new diagnostic targets are incorporated through only lightweight task-specific heads, enabling a “unified-model, multi-disease” paradigm that approaches plug-and-play clinical deployment.

Several limitations warrant consideration. First, the use of unconditional diffusion models limits the generative potential of the framework; in the context of musculoskeletal MRI, controlled diffusion strategies could enable targeted data augmentation and help alleviate class imbalance and data scarcity. Second, the current study focuses primarily on joint-related musculoskeletal disorders assessed via MRI. Extending OrthoDiffusion to multi-modal inputs and additional anatomical sites, alongside more diverse datasets covering a broader spectrum of clinical tasks, represents an important future direction. Third, while the model demonstrates strong performance across multiple centers and magnetic field strengths, it has not yet been validated in real-time clinical workflows. Incorporating human–AI collaboration and prospective evaluations will be essential to assess its usability, reliability, and ultimate impact on diagnostic decision-making. Despite these limitations, OrthoDiffusion represents a meaningful advance toward a unified, diffusion-based paradigm for musculoskeletal MRI.

\section{Methods}\label{sec:methods}

\subsection{Ethics statement}
This retrospective multi-center study was conducted in accordance with the Declaration of Helsinki and was approved by the Peking University Third Hospital Medical Science Research Ethics Committee (Project Number IRB00006761-M2024043) and other participating centers. Due to the retrospective nature of the study and the anonymization of all patient data, the requirement for informed consent was waived. 

\subsection{Dataset}
\label{sec:method_dataset}
\subsubsection*{Study Cohort}
We curated a large-scale musculoskeletal MRI dataset comprising over 30,000 examinations from nine independent clinical centers. The cohort included patients presenting with joint-related symptoms who underwent MRI scans. To ensure data quality and consistency across datasets, we applied strict exclusion criteria: (1) history of prior surgical intervention or hardware implantation in the target joint; (2) presence of acute complex fractures or tumors causing significant anatomical distortion; (3) severe motion or metal artifacts preventing diagnostic interpretation; and (4) incomplete imaging sequences. 

\begin{sidewaystable}
\centering
\setlength{\tabcolsep}{1.5pt}
\caption{Overview of datasets used for knee-only diffusion pre-training and cross-joint downstream evaluation.}
\label{tab:dataset_overview}
\begin{tabular}{l l l l c c c c}
\hline
Task
& Joint
& Centers 
& No. of MRI 
& Abnormalities / Structures 
& Age (years) 
& Sex (M/F) 
& Field strength (1.5T / 3T) \\
\hline

Diffusion pretraining
& Knee
& A--C 
& 15,948 
& -- 
& 33.0 (3--87) 
& 65.3\% / 34.7\% 
& 28.8\% / 71.2\% \\

Knee diagnostic
& Knee
& A--G 
& 10,940
& 8 knee abnormalities 
& 33.4 (6--87)
& 65.5\% /34.5\%
& 24.9\% / 75.1\%\\

Knee anatomical segmentation 
& Knee
& A--B 
& 1,006 
& 11 anatomical structures 
& 42.4 (6--80) 
& 55.9\% / 44.1\% 
& 31.5\% / 68.5\% \\

Ankle diagnostic
& Ankle
& A--C, H 
& 2,562 
& 4 ankle abnormalities 
& 33.4 (6--80) 
& 49.3\% / 50.7\% 
& 7.0\% / 93.0\% \\

Shoulder diagnostic
& Shoulder
& A--C, H--I 
& 8,957 
& 7 shoulder abnormalities 
& 49.6 (9--93) 
& 52.1\% / 47.9\% 
& 20.3\% / 79.7\% \\
\hline
\end{tabular}
\end{sidewaystable}

\subsubsection*{Dataset Composition}
The dataset was organized into three distinct subsets to support model pretraining, downstream tasks, and cross-anatomy generalization. Demographic details and acquisition parameters of datasets are provided in Table~\ref{tab:dataset_overview} and Supplementary Table~\ref{tab:sag_knee_seg_stats}-\ref{tab:shoulder_mri_stats}.

\textbf{Pretraining Dataset.} To build robust feature representations, we utilized 15,948 unlabeled knee MRI examinations collected from Centers A–C between December 2016 and June 2023. These images, acquired at magnetic field strengths of 1.5T and 3.0T using routine sagittal, coronal, and axial proton density-weighted (PDW) sequences, were used exclusively for self-supervised diffusion model pretraining. 

\textbf{Knee Diagnostic and Segmentation Datasets.} For the classification task, we compiled a labeled dataset of 10,940 knee MRI scans from eight institutions (Centers A–G) between December 2016 and March 2025. This subset features high heterogeneity, utilizing scanners from multiple vendors (GE Healthcare, Siemens, Philips, and UIH) to test robustness in real-world scenarios. Additionally, we also constructed a dedicated knee segmentation dataset consisting of 1,006 knee MRI examinations collected from two institutions (Centers A and B) between July 2023 and February 2024. Unlike the classification subset, this cohort focused on pixel-wise anatomical delineation using sagittal and coronal PDW sequences.

\textbf{Cross-Anatomy Generalization Datasets.} To assess the transferability of the learned representations to other joints, we incorporated independent datasets for ankle and shoulder MRI. The ankle cohort comprised 2,562 examinations collected from three institutions (Centers A–C and H) between June 2024 and June 2025. The shoulder cohort included 8,957 examinations from four institutions (Centers A–C, H and I) collected between January 2023 and June 2025. Both datasets included sagittal, coronal, and axial PDW sequences acquired at 1.5T or 3.0T. These datasets were not seen during pretraining and served exclusively to validate the model’s ability to generalize to anatomical structures and pathologies.

\subsubsection*{Establishment of Ground Truth}
Ground truth annotations were generated through rigorous, multi-stage, task-specific protocols. For classification datasets, the reference standard was determined based on clinical intervention status. For patients who underwent surgery, intraoperative findings served as the definitive gold standard. For non-surgical cases, labels were independently generated by three board-certified specialists in sports medicine, each with over ten years of experience. These reviewers performed blinded assessments of multi-view sequences without access to original reports. Any diagnostic discrepancies were resolved through consensus discussion among the three experts to establish the final gold standard. For segmentation dataset, we employed a hierarchical multi-stage annotation strategy. Initial segmentation masks for 11 distinct anatomical structures were manually delineated by six junior radiologists. All initial annotations were subsequently audited by three senior radiologists, each possessing over seven years of experience. Annotations deemed inaccurate were returned to the junior annotators for revision. For complex or ambiguous cases, a consensus ground truth mask was generated through consultation among three senior experts.

\subsubsection*{Data preprocessing}
During both training and testing, MRI scans are standardized by extracting a contiguous stack of 16 central slices from each PDW sequence and resizing each slice to a spatial resolution of $256 \times 256$. Volumes are normalized using min–max scaling, followed by linear intensity mapping to the range $[-1,1]$.

\subsection{Model Overview}
\label{sec:methods/model_overview}

OrthoDiffusion is an orientation-aware diffusion framework that learns robust and transferable anatomical representations from musculoskeletal MRI (Fig.~\ref{fig:overview} (b)-(d)). OrthoDiffusion pretrains three independent diffusion backbones-one for each orientation-to capture view-specific structural priors through large-scale self-supervised denoising.

For downstream task analysis, OrthoDiffusion extracts intermediate features from the bottleneck layer of each pretrained diffusion model at a selected diffusion timestep. These features are then processed through the pooling operator and subsequently combined using the fusion strategy. 

OrthoDiffusion adopts a unified \textbf{two-stage} training protocol for \textbf{classification tasks}. In Stage \textbf{I}, diffusion backbones are used to generate orientation-specific feature maps: for \textit{linear probing}, the backbone is fully frozen and only the pooling module is optimized; for \textit{fine-tuning}, the pooling module and the encoder/bottleneck layers of the diffusion 3D U-Net are jointly updated. After convergence of Stage I, all feature-extracting components are frozen. In Stage \textbf{II}, only the fusion module is optimized, ensuring that cross-orientation integration is learned over stable, task-adapted representations without interference from upstream gradient updates.

For \textbf{anatomical segmentation tasks}, we adopt a minimal fine-tuning strategy in which the encoder and bottleneck layers of the diffusion backbone are jointly optimized together with a shallow segmentation head. No additional pooling or multi-orientation fusion modules are introduced in the segmentation pipeline. 

All representations are passed to task-specific heads, including a single-layer linear classifier trained with binary cross-entropy (BCE) for multi-label diagnosis and a lightweight segmentation head optimized with cross-entropy and soft Dice losses for anatomical segmentation.

% \begin{figure*}[t]
%     \centering
%     \includegraphics[width=\textwidth]{figures/pipeline.pdf}
%     \caption{
%     Overview of the C-UNIDIFF framework.
%     (a) Unconditional 3D diffusion pretraining using a 3D U-Net noise predictor.
%     (b) Feature extraction from intermediate diffusion representations at a selected timestep and bottleneck block, followed by pooling and multi-label classification.
%     (c) Stage \textbf{I}: feature-level and label-level fusion strategies for integrating sagittal, coronal, and axial representations.
%     (d) Stage \textbf{II}: segmentation pipeline using diffusion representations and a lightweight segmentation head.
%     (e) A unified diffusion backbone supporting multiple clinical tasks, including knee, ankle, shoulder diagnosis, and knee anatomical segmentation.
%     (f) Multi-orientation fusion during validation and testing for diagnostic classification.
%     (g) Anatomical segmentation using diffusion features without multi-orientation fusion.
%     Trainable and frozen modules are indicated by flame and snowflake icons, respectively.
%     }
%     \label{fig:pipeline}
% \end{figure*}

\subsection{Orientation-Specific Diffusion Backbone Pretraining}
\label{sec:methods/pretrained_diffusion_model}

Because sagittal, coronal, and axial views provide complementary anatomical information in musculoskeletal MRI, OrthoDiffusion trained three independent unconditional 3D diffusion models, each corresponding to a single MRI orientation. This design enables each model to capture the 3D distributional characteristics specific to its corresponding view, yielding anatomically coherent and orientation-aware latent representations for downstream tasks~\cite{ddae2023, Baranchuk2021LabelEfficientSS}.

Each diffusion model adopts a 3D U-Net denoiser to parameterize the reverse diffusion process, following established volumetric diffusion architectures~\cite{med-ddpm2024}. This architecture captures multi-scale spatial context and supports high-fidelity reconstruction of anatomical structures within each orientation. 

Diffusion models are capable of approximating highly complex data distributions $p(\mathbf{x})$~\cite{Ho2020DenoisingDP}. Specifically, Gaussian noise is progressively added to a 3D input volume $\mathbf{x}_0$ over $T$ diffusion timesteps according to \(\mathbf{x}_{t}=\sqrt{\alpha_{t}} \mathbf{x}_{0}+\sqrt{1-\alpha_{t}} \bm{\varepsilon}\) ,with \(t=1, \cdots, T, \bm{\varepsilon} \sim N(0, \mathbf{I})\), where \(\alpha_{t}\) controls the noise level at timestep $t$. The model is trained to learn the corresponding reverse denoising process of this fixed-length Markov chain. The learning objective is formulated as
\begin{equation}
\label{eq: diffusion}
\boldsymbol{\theta}^{*}
=
\arg\min_{\boldsymbol{\theta}}
\; \mathbb{E}_{\mathbf{x},\, \boldsymbol{\varepsilon},\, t}
\left[
\left\|
\boldsymbol{\varepsilon}
-
\boldsymbol{\varepsilon}_{\boldsymbol{\theta}}\!\left(\mathbf{x}_{t}, t\right)
\right\|_{2}^{2}
\right].
\end{equation}

where \(\bm{\varepsilon}_{\bm{\theta}}(\cdot)\)  denotes the neural network that predicts the injected noise.

\subsection{Extraction of Intermediate Diffusion Representations}
\label{sec:methods/extract representation pooling}

For each MRI orientation, OrthoDiffusion extracts intermediate
feature maps from the pretrained diffusion denoiser $\bm{\epsilon}_{\bm{\theta}}(\mathbf{x}_t, t)$. These intermediate activations encode multi-scale, orientation-specific anatomical representations for downstream analysis.

To transform high-dimensional diffusion feature maps into compact and discriminative embeddings suitable for classification tasks, OrthoDiffusion evaluates three pooling strategies. \textit{Global Average Pooling (GAP)} produces a holistic descriptor by averaging features across all spatial locations. \textit{Global–Local Pooling (GLP)} augments the global summary with a locally attended feature, enabling the retention of region-specific structural information. \textit{Self-Attention Pooling (SAP)} applies multi-head self-attention to adaptively weight informative spatial locations and fuses the resulting representation with a global descriptor, enabling the modeling of long-range dependencies.

Unless otherwise stated, SAP is adopted as the default pooling strategy in all experiments, due to its superior and stable performance observed in ablation studies (Supplementary Table~\ref{tab:linearprobing-pooling-aucroc} and~\ref{tab:linearprobing-pooling-multilabel}). 

\subsection{Task-Specific Selection of Diffusion Timesteps and Bottleneck Blocks}
\label{sec:methods/timesteps}

OrthoDiffusion evaluates diffusion representations across sagittal, coronal, and axial views at multiple diffusion timesteps and bottleneck blocks. The optimal configuration is selected on the validation set and fixed for all held-out test cohorts to avoid data leakage.

For each task, MRI orientation and training setting (linear probing or fine-tuning), OrthoDiffusion selects the best-performing "Timestep-Block" combination, which are summarized in Supplementary Table~\ref{tab:timestep_block_selection}.

\subsection{Integrating Complementary Information Across MRI Orientations}
\label{sec:methods/fusion}
 
To integrate complementary information across the MRI three orientations, OrthoDiffusion assesses both feature-level and label-level fusion strategies for combining orientation-specific representations. 

At the feature level, OrthoDiffusion considers several commonly used approaches for merging
orientation-specific embeddings prior to downstream tasks. The simplest method is direct concatenation, which retains all orientation-wise information without additional parameters.  
OrthoDiffusion further considers linear-projection fusion, where each orientation-specific embedding is mapped into a shared latent space and merged by element-wise addition or concatenation, enabling lightweight learnable alignment across orientations. Finally, OrthoDiffusion explores a cross-attention mechanism that enables the three orientation features to exchange information; each embedding attends to structural cues present in the others, and the refined embeddings are then concatenated for downstream prediction.

Complementing feature-level fusion, OrthoDiffusion introduces a label-level strategy
termed \textit{Multi-plane Adaptive Expert Fusion} (MPAE), designed primarily to improve interpretability across anatomical orientations. MPAE takes the logits produced by the three orientation-specific classifiers and uses a compact gating network to generate patient- and label-specific fusion weights. After softmax normalization, these weights determine how the three logits are combined. This design highlights the most informative orientation for each diagnostic label and yields clinically interpretable weighting patterns that mirror how radiologists integrate sagittal, coronal, and axial cues.

Unless otherwise stated, simple concatenation is used as the default fusion strategy in all subsequent experiments, based on ablation analyses reported in the Supplementary Table~\ref{tab:linearprobing-fusion-aucroc} and~\ref{tab:linearprobing-fusion-multilabel}.  

\subsection{Multimodal Fusion with Electronic Health Records}

To further exploit multimodal information, OrthoDiffusion incorporates Electronic Health Records (EHR) associated with each subject through a unique hospitalization identifier. The EHR variables include demographic attributes (age, sex, height, weight), patient type (athlete vs.\ non-athlete), and categorical injury-inducing events.

Continuous variables are standardized, while categorical attributes are encoded using one-hot vectors or learnable embeddings, with missing values handled through explicit \texttt{UNK} categories to avoid cohort exclusion.

MRI and EHR information are integrated via a late-fusion strategy at the logit level (Fig.~\ref{fig:overview} (e)). Specifically, independent prediction heads are applied to MRI features and EHR features, and the resulting logits are combined through learnable, label-wise fusion weights. This design preserves the discriminative power of imaging features while enabling selective utilization of complementary clinical signals. Detailed implementation is provided in the Supplementary Materials.

\subsection{Evaluation and statistical analysis}

Task performance was evaluated using standard metrics for classification and segmentation, including the area under the receiver operating characteristic curve (AUROC), average precision (AP), class-wise precision, recall, and F1-score (CP/CR/CF1), overall precision, recall, and F1-score (OP/OR/OF1), and the Dice Similarity Coefficient (DSC).

For multi-label classification, AUROC was computed separately for each disease category and macro-averaged across tasks. AP summarizes the precision–recall curve, while CP/CR/CF1 and OP/OR/OF1 quantify class-wise and overall classification performance, respectively. Segmentation performance was assessed using DSC to measure the spatial overlap between predicted masks and ground-truth annotations.

Statistical significance between OrthoDiffusion and baseline models was assessed using paired non-parametric permutation tests on per-task performance differences. Across all evaluated tasks, permutation tests indicated statistically significant and consistent performance improvements ($ p < 0.001$).

\clearpage
\backmatter

% \bmhead{Supplementary information}

% If your article has accompanying supplementary file/s please state so here. 

% Authors reporting data from electrophoretic gels and blots should supply the full unprocessed scans for key as part of their Supplementary information. This may be requested by the editorial team/s if it is missing.

% Please refer to Journal-level guidance for any specific requirements.

% \bmhead{Acknowledgements}

% Acknowledgements are not compulsory. Where included they should be brief. Grant or contribution numbers may be acknowledged.

% Please refer to Journal-level guidance for any specific requirements.
\section*{Acknowledgment}
This work was funded by the National Natural Science Foundation of China (Grant number 82441025), and Beijing Municipal Natural Science Foundation (Grant number L242104). 

\section*{Data Availability}
The datasets are not publicly available due to privacy concerns, but are partially available from the corresponding author on reasonable request.

\section*{Code Availability}
All code is available via GitHub at \url{https://github.com/lt-0123/OrthoDiffusion}.

\section*{Author Contributions}
T.L., L.X., Z.Y., W.X., H.Y., J.Y., and D.W. conceived and designed the study.  L.X., Z.Y., S.L., J.L., and J.L. acquired, organized, and verified the raw data. T.L. developed the methodology, performed the technical implementation, and conducted the results analysis. T.L., L.X., Z.Y., D.W., J.Y. discussed the results and provided critical comments on the paper. The study was supervised by D. J, J.Y. and D. W. All authors contributed to the drafting and revising of the manuscript and approved the final version.

\section*{Competing Interests}
The authors declare no competing interests.

% \section*{Declarations}

% Some journals require declarations to be submitted in a standardised format. Please check the Instructions for Authors of the journal to which you are submitting to see if you need to complete this section. If yes, your manuscript must contain the following sections under the heading `Declarations':

% \begin{itemize}
% \item Funding
% \item Conflict of interest/Competing interests (check journal-specific guidelines for which heading to use)
% \item Ethics approval and consent to participate
% \item Consent for publication
% \item Data availability 
% \item Materials availability
% \item Code availability 
% \item Author contribution
% \end{itemize}

% \noindent
% If any of the sections are not relevant to your manuscript, please include the heading and write `Not applicable' for that section. 

%%===================================================%%
%% For presentation purpose, we have included        %%
%% \bigskip command. Please ignore this.             %%
%%===================================================%%
\bigskip
\clearpage

\begin{appendices}
\section{}

\setcounter{figure}{0}
\renewcommand{\thefigure}{\arabic{figure}}
\captionsetup[figure]{name=Extended Data Fig.}

\begin{figure}[h]
    \centering
    % 第一张图
    \begin{subfigure}[b]{0.32\textwidth}
        \centering
        \includegraphics[width=\linewidth]{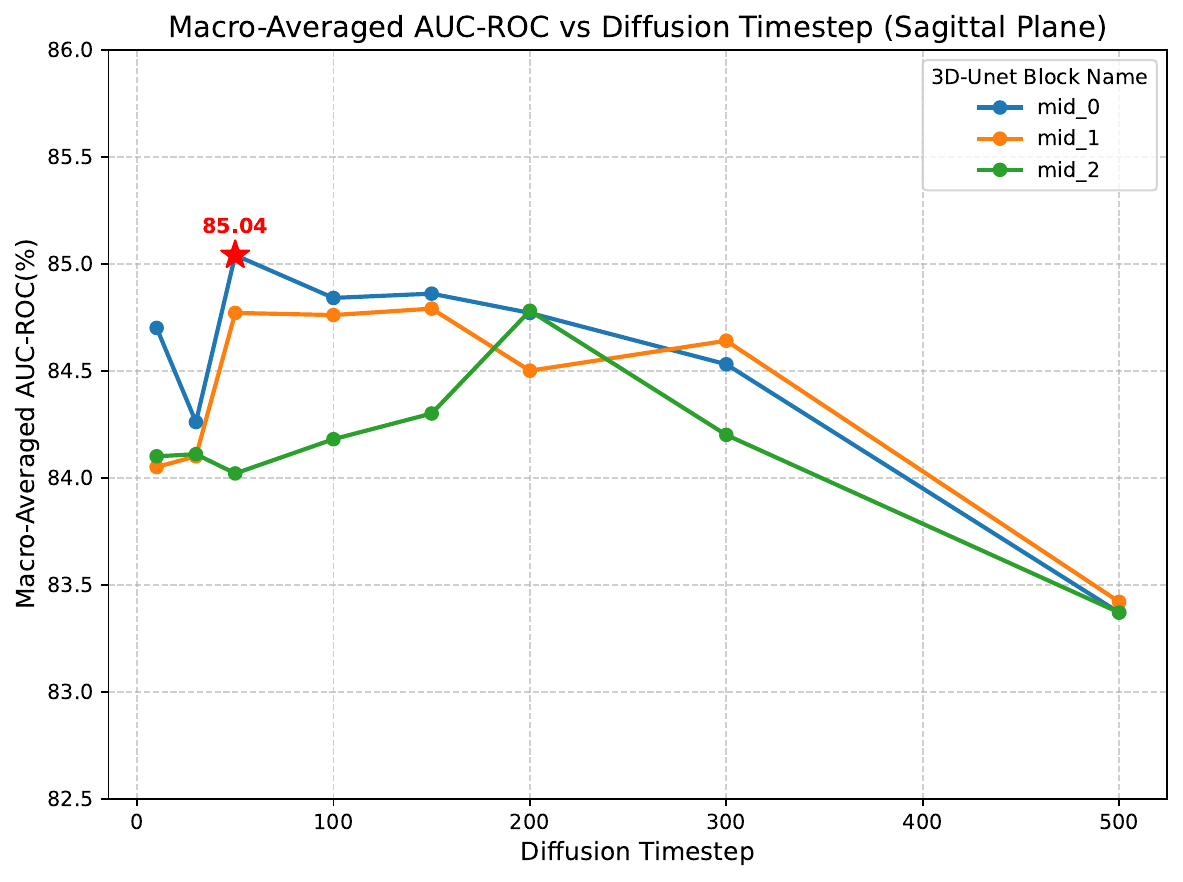}
        \caption{Sagittal Plane}
        \label{fig:sagittal}
    \end{subfigure}
    \hfill
    % 第二张图
    \begin{subfigure}[b]{0.32\textwidth}
        \centering
        \includegraphics[width=\linewidth]{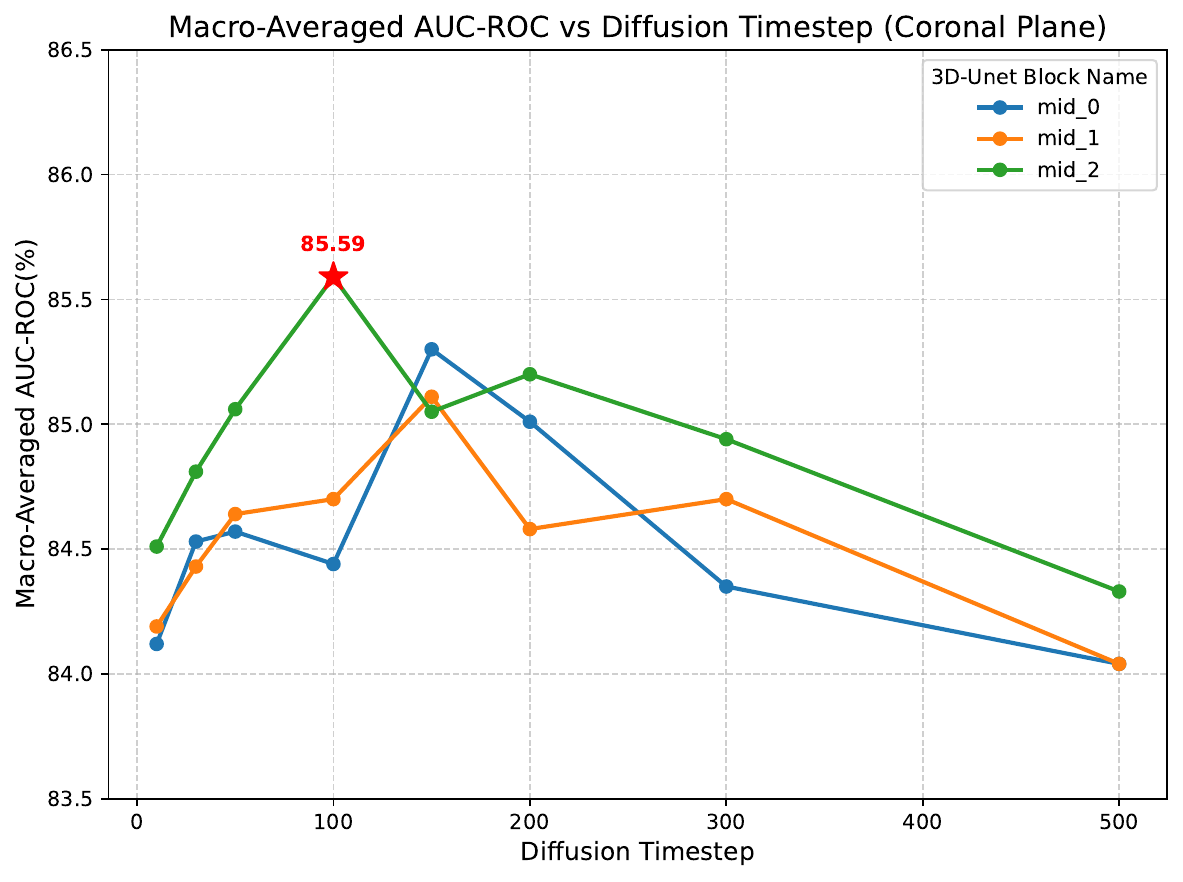}
        \caption{Coronal Plane}
        \label{fig:coronal}
    \end{subfigure}
    \hfill
    % 第三张图
    \begin{subfigure}[b]{0.32\textwidth}
        \centering
        \includegraphics[width=\linewidth]{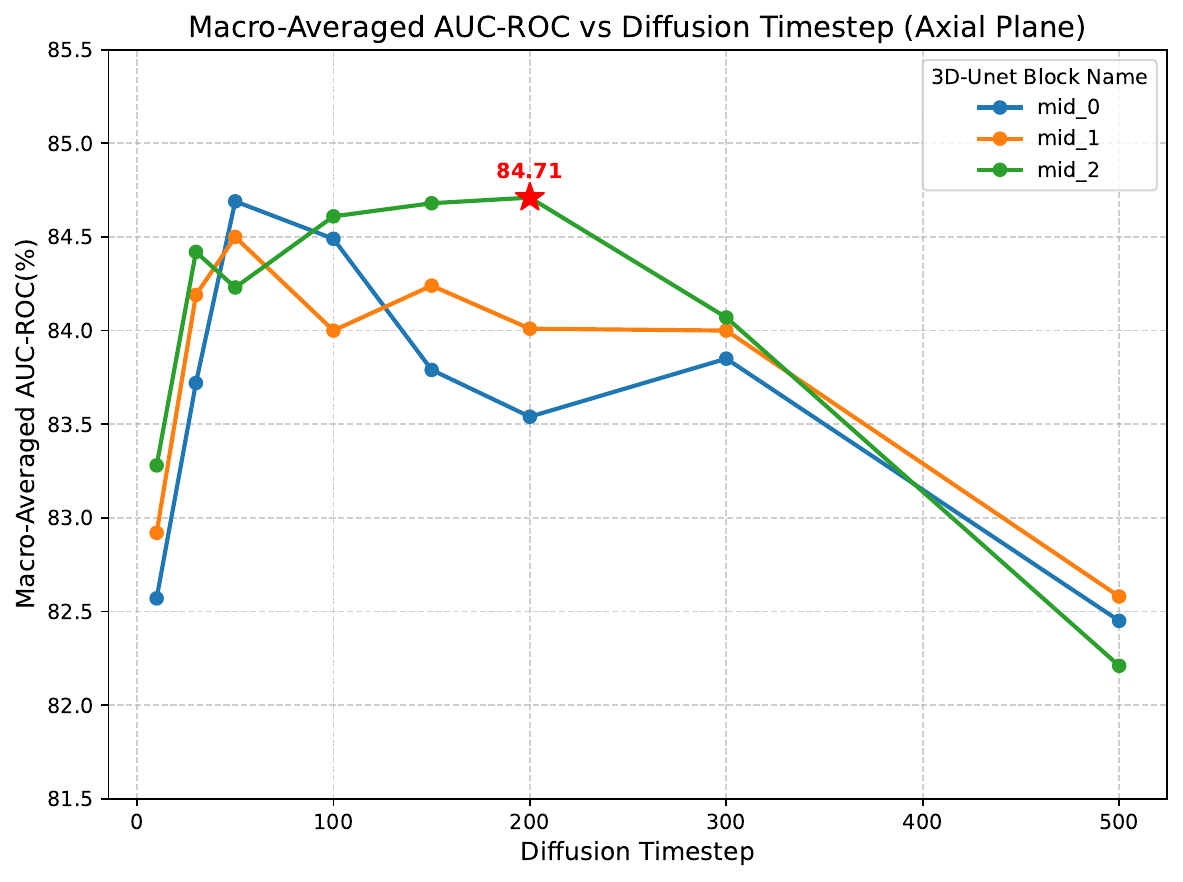}
        \caption{Axial Plane}
        \label{fig:axial}
    \end{subfigure}

    \caption{Macro-averaged AUROC of the eight-label \textbf{knee injury prediction} task under linear probing on the Center A+B+C test set, based on SAP-processed features extracted from 3D-UNet blocks at different diffusion timesteps across MRI planes.}
    \label{fig:aucroc_comparison}
\end{figure}
\newpage

\section{More Detail about Dataset}\label{app:dataset}
Detailed summary statistics for all datasets used in this study are provided in Supplementary Table~\ref{tab:sag_knee_seg_stats}-\ref{tab:shoulder_mri_stats}. These tables report the composition of training, validation, and test cohorts across anatomical segmentation and diagnostic classification tasks, including scanner manufacturers, magnetic field strengths, acquisition parameters, and patient demographics.

\section{More Detail about Methods}
\subsection{Pooling Strategies for Representation Extraction}
\label{sec:appendix_pooling}
For each plane \(o \in \mathcal{O}=  \{\text{sagittal},\,\text{coronal},\,\text{axial}\}\), OrthoDiffusion extracts intermediate diffusion representations \(\mathbf{v}_{t,b}^{(o)} \in \mathbb{R}^{C \times D' \times H' \times W'}\) from the pre-trained diffusion model \(\bm{\epsilon}_{\bm{\theta}}(\mathbf{x}_t, t)\). The MRI sequence input is \(\mathbf{x}_0 \in \mathbb{R}^{1 \times D \times H \times W}\), and is gradually perturbed into \(\mathbf{x}_t\) at timestep \(t\). Here, \(b\) indexes 3D U-Net blocks, \(C\) the channel dimension, and \(D'\!,H'\!,W'\) the downsampled spatial size of the corresponding feature map. OrthoDiffusion evaluates three pooling strategies for deriving compact representations $f_{t,b}^{(o)}$:

\begin{enumerate}
    \item \textbf{Global average pooling (GAP)}:  
    The most straightforward approach is to apply global average pooling over the spatial dimensions:
    \[
    f_{\text{global}}^{(o)} = \frac{1}{D'H'W'} 
    \sum_{d=1}^{D'} \sum_{h=1}^{H'} \sum_{w=1}^{W'} \mathbf{v}_{t,b}^{(o)}[:,d,h,w]
    \;\in\; \mathbb{R}^C,
    \]
    yielding $f_{t,b}^{(o)} = f_{\text{global}}^{(o)}$.

    \item \textbf{Global–local pooling (GLP)}:  
    In addition to the global descriptor from GAP, a local attention mechanism highlights spatially discriminative regions. 
    Let $\mathbf{v}_{t,b,i}^{(o)} \in \mathbb{R}^C$ denote the feature vector at spatial index $i \in \{1,\ldots,N\}$ with $N = D'H'W'$. 
    Attention weights are computed as
    \[
    \alpha_i = \frac{\exp\!\big(g(v_{t,b,i}^{(o)})\big)}{\sum_{j=1}^N \exp\!\big(g(v_{t,b,j}^{(o)})\big)}, 
    \quad 
    f_{\text{local}}^{(o)} = \sum_{i=1}^N \alpha_i\, v_{t,b,i}^{(o)},
    \]
    where $g(\cdot)$ is a two-layer MLP with ReLU activation. 
    The global and local descriptors are then concatenated to form the final representation:
    \[
    f_{t,b}^{(o)} = \big[ f_{\text{global}}^{(o)} \;\|\; f_{\text{local}}^{(o)} \big] \in \mathbb{R}^{2C}.
    \]
    
    \item \textbf{Self-attention pooling (SAP)}:  
    OrthoDiffusion interprets the spatial features $\{\mathbf{v}_{t,b,i}^{(o)}\}_{i=1}^N$ as a sequence of tokens and arrange them into
    \[
    X = [\mathbf{v}_{t,b,1}^{(o)}, \ldots, \mathbf{v}_{t,b,N}^{(o)}] \in \mathbb{R}^{N \times C}.
    \]
    Each token is linearly projected into queries, keys, and values:
    \[
    Q = X W_Q,\quad K = X W_K,\quad V = X W_V,
    \quad W_Q, W_K, W_V \in \mathbb{R}^{C \times C}.
    \]
    
    We split the feature dimension into $h$ heads of size $d = C/h$, producing
    \[
    Q = [Q_1 \| \cdots \| Q_h],\quad 
    K = [K_1 \| \cdots \| K_h],\quad
    V = [V_1 \| \cdots \| V_h],\quad
    Q_j, K_j, V_j \in \mathbb{R}^{N \times d}.
    \]
    
    Scaled dot-product attention is computed independently for each head, where $j=1,2,\cdots, h$:
    \[
    \text{head}_j =
    \text{softmax}\!\left(\frac{Q_j K_j^\top}{\sqrt{d}}\right)V_j,
    \quad \text{head}_j \in \mathbb{R}^{N \times d}.
    \]
    
    The multi-head output is formed as
    \[
    \text{MHA}(X) = \text{Concat}(\text{head}_1,\ldots,\text{head}_h) W_O \in \mathbb{R}^{N \times C},
    \quad W_O \in \mathbb{R}^{C \times C}.
    \]
    Let $\text{MHA}(X)_i \in \mathbb{R}^C$ denote the $i$-th token embedding (the $i$-th row of $\text{MHA}(X)$). 
    A compact representation is then obtained by average pooling over tokens:
    \[
    f_{\text{attn}}^{(o)} = \frac{1}{N} \sum_{i=1}^N \text{MHA}(X)_i
    \in \mathbb{R}^C.
    \]
    
    Finally,
    \[
    f_{t,b}^{(o)} = \big[ f_{\text{global}}^{(o)} \;\|\; f_{\text{attn}}^{(o)} \big]
    \in \mathbb{R}^{2C}.
    \]

\end{enumerate}

GLP emphasizes spatially localized discriminative structures through learned attention, while SAP captures long-range dependencies via self-attention. Both mechanisms complement the holistic context provided by GAP.

For brevity, the pooled representation \( f_{t,b}^{(o)} \) will be referred to as \( \mathbf{F}^{(o)} \) hereafter.

\subsection{Fusion Strategies Across MRI Orientations}
\label{sec:appendix fusion strategies}

For each \( o \in \mathcal{O} \), let
\(
\mathbf{F}^{(o)} \in \mathbb{R}^{C'}
\)
denote the orientation-specific feature representation extracted by the corresponding diffusion backbone, where \( C' = C \) for GAP and \( C' = 2C \) for GLP and SAP pooling.
OrthoDiffusion compares four fusion strategies for integrating these representations:

\subsubsection*{(1) Simple concatenation.}
The fused representation is obtained by channel-wise concatenation:
\[
\mathbf{F}_{\mathrm{fuse}}
    =
    \bigl[
        \mathbf{F}^{(\mathrm{sag})}
        \,\|\,
        \mathbf{F}^{(\mathrm{cor})}
        \,\|\,
        \mathbf{F}^{(\mathrm{ax})}
    \bigr]
    \in \mathbb{R}^{3C'}.
\]

\subsubsection*{(2) Linear fusion.}
To allow fusion in a shared latent space, each feature vector
\(
\mathbf{F}^{(o)} \in \mathbb{R}^{C'}
\)
is linearly projected into a common embedding dimension \(E\):
\[
\mathbf{H}^{(o)} = W^{(o)}\, \mathbf{F}^{(o)} \in \mathbb{R}^{E}, \qquad o \in \mathcal{O}.
\]
Fusion is then performed either by element-wise addition \textbf{(Linear addition)}:
\[
\mathbf{F}_{\mathrm{fuse}}
    =
    \mathbf{H}^{(\mathrm{sag})}
    +
    \mathbf{H}^{(\mathrm{cor})}
    +
    \mathbf{H}^{(\mathrm{ax})}
    \in \mathbb{R}^{E},
\]
or by channel-wise concatenation \textbf{(Linear concatenation)}:
\[
\mathbf{F}_{\mathrm{fuse}}
    =
    \bigl[
        \mathbf{H}^{(\mathrm{sag})}
        \,\|\,
        \mathbf{H}^{(\mathrm{cor})}
        \,\|\,
        \mathbf{H}^{(\mathrm{ax})}
    \bigr]
    \in \mathbb{R}^{3E}.
\]

\subsubsection*{(3) Cross-attention fusion.}
To enable information exchange across poses, all features are first projected into the same embedding:
\[
\mathbf{H}^{(o)} = W^{(o)}\,\mathbf{F}^{(o)} \in \mathbb{R}^{E}, \quad o\in\mathcal{O}.
\]
Cross-attention is used to enrich each orientation with contextual cues from the others, e.g.,
\[
\widetilde{\mathbf{H}}^{(\mathrm{sag})}
    =
    \mathrm{MHA}\bigl(
        \mathbf{H}^{(\mathrm{sag})},\,
        \mathbf{H}^{(\mathrm{cor})},\,
        \mathbf{H}^{(\mathrm{cor})}
    \bigr),
\]
and the refined embeddings are obtained via residual connections and layer normalization:
\[
\mathbf{H}'^{(o)}
    =
    \mathrm{LayerNorm}
    \Bigl(
        \mathbf{H}^{(o)} + \widetilde{\mathbf{H}}^{(o)}
    \Bigr),
    \quad o\in\mathcal{O}.
\]
Finally, feature fusion is performed by concatenation:
\[
\mathbf{F}_{\mathrm{fuse}}
    =
    \bigl[
        \mathbf{H}'^{(\mathrm{sag})}
        \,\|\,
        \mathbf{H}'^{(\mathrm{cor})}
        \,\|\,
        \mathbf{H}'^{(\mathrm{ax})}
    \bigr]
    \in \mathbb{R}^{3E}.
\]
\subsubsection*{(4) MPAE fusion.}

For each anatomical orientation \( o \in \mathcal{O} \), an independent multi-label classifier outputs
\[
\mathbf{z}^{(o)} = \bigl(z^{(o)}_1,\ldots,z^{(o)}_K\bigr) \in \mathbb{R}^K,
\]
where \( z^{(o)}_k \) denotes the predicted logits for disease category \( k \), and \( K \) is the number of musculoskeletal diseases.

For each label \( k \), we form the orientation-wise logits vector
\[
\mathbf{z}_k = \bigl(z^{(\mathrm{sag})}_k,\, z^{(\mathrm{cor})}_k,\, z^{(\mathrm{ax})}_k \bigr)^\top \in \mathbb{R}^3.
\]
An adaptive gating network \( g(\cdot) \) produces unnormalized scores \( \mathbf{s}_k = g(\mathbf{z}_k) \in \mathbb{R}^3 \), and the fusion weights are obtained by
\[
\boldsymbol{\alpha}_k = \mathrm{softmax}(\mathbf{s}_k),
\qquad
\sum_{o\in\mathcal{O}} \alpha^{(o)}_k = 1.
\]

Fusion is performed in logit space:

\begin{equation}
\label{eq: MAE}
    \hat{z}_k = \sum_{o\in\mathcal{O}} \alpha^{(o)}_k\, z^{(o)}_k,
    \qquad
    \hat{y}_k = \sigma(\hat{z}_k).
\end{equation}

This label-specific, orientation-aware aggregation allows the model to emphasize the most informative anatomical plane for each disorder, mirroring the way radiologists integrate complementary cues across MRI views.

\subsection{Integration of EHR data}
\label{sec:appendix intergrate clinical}

\subsubsection*{EHR variables.}
To assess whether structured clinical information provides complementary
predictive value beyond MRI-derived representations, OrthoDiffusion incorporates Electronic
Health Records (EHR) linked to each subject via a unique hospitalization
identifier. The variables include:

\begin{itemize}
    \item \textbf{Patient type}: athlete vs.\ non-athlete.
    \item \textbf{Sex}: male vs.\ female.
    \item \textbf{Anthropometric measurements}:
    \begin{itemize}
        \item age (years),
        \item height (cm),
        \item weight (kg).
    \end{itemize}
    \item \textbf{Injury-inducing event}, categorized into eight mutually
    exclusive classes:
    \begin{enumerate}
        \item overuse injury,
        \item traffic accident,
        \item military or police duty accident,
        \item daily-life or production accident,
        \item sports-related accident,
        \item spontaneous (no identifiable trigger),
        \item actor or rehearsal accident,
        \item other.
    \end{enumerate}
\end{itemize}

All records are extracted from the hospital information system and aligned with
MRI features via the unique patient identifier. Missing or unavailable entries
are preserved by assigning an \texttt{UNK} (unknown) category rather than
discarding samples, ensuring consistent cohort size and preventing bias
introduced by sample exclusion.

\subsubsection*{EHR Feature Encoding and Normalization.}
Continuous variables (\textbf{age, height, weight}) are standardized using a
\textit{z}-score transform. Missing numeric values are imputed with zero after
normalization. Categorical variables (\textbf{sex} and \textbf{patient type})
are converted to fixed-length one-hot vectors, including an \texttt{UNK}
category to accommodate missing or ambiguous entries.

The \textbf{injury–inducing event} is modeled as a learnable discrete token:
each event is mapped to an integer index and embedded into an 8-dimensional
vector via a trainable lookup table. Records without a valid entry receive an
\texttt{UNK} token automatically.

For each subject, the resulting structured EHR representation is
\[
\mathbf{F}_{\mathrm{EHR}}
=
\big[
\underbrace{\mathbf{x}^{(\mathrm{cont})}}_{\in\mathbb{R}^3}
\;\|\;
\underbrace{\mathbf{x}^{(\mathrm{sex})}}_{\in\mathbb{R}^3}
\;\|\;
\underbrace{\mathbf{x}^{(\mathrm{ptype})}}_{\in\mathbb{R}^3}
\;\|\;
\underbrace{\mathbf{e}}_{\in\mathbb{R}^{8}}
\big],
\]
where \(\mathbf{x}^{(\mathrm{cont})}\) denotes standardized continuous
measurements, \(\mathbf{x}^{(\mathrm{sex})}\) and
\(\mathbf{x}^{(\mathrm{ptype})}\) are one-hot vectors, and \(\mathbf{e}\) is the
learned event embedding. This design enables structured clinical attributes to
be seamlessly integrated with MRI representations in downstream classification.

\subsubsection*{MRI–EHR Fusion Strategy.}
To quantify the contribution of structured clinical attributes while avoiding scale imbalance between modalities, OrthoDiffusion adopts a \emph{late–fusion} architecture operating
at the \textbf{logit level}. Given a 1536-dimensional MRI representation
\(\mathbf{F}_{\mathrm{MRI}}\) (after Simple Concat fusion) and an EHR
representation \(\mathbf{F}_{\mathrm{EHR}}\), two independent multilabel heads
produce
\[
\mathbf{z}_{\mathrm{MRI}},\;
\mathbf{z}_{\mathrm{EHR}} \in \mathbb{R}^{K},
\]
where \(K\) denotes the number of disease categories.

Fusion is parameterized by a learnable, per-label coefficient
\(\boldsymbol{\gamma}\in(0,1)^{K}\):
\[
\hat{\mathbf{z}}
=
\boldsymbol{\gamma}\odot\mathbf{z}_{\mathrm{MRI}}
+
(1-\boldsymbol{\gamma})\odot\mathbf{z}_{\mathrm{EHR}},
\]
with \(\odot\) representing element-wise multiplication.
Fusion weights are given by
\[
\boldsymbol{\gamma} = \sigma(\mathbf{w}), \quad \mathbf{w}\in\mathbb{R}^{K},
\]
where \(\sigma(\cdot)\) is the element-wise logistic sigmoid.

The fused logits yield final probabilities
\[
\hat{\mathbf{y}} = \sigma(\hat{\mathbf{z}})\in[0,1]^K.
\]

This formulation (i) preserves the discriminative capacity of MRI-derived
representations, (ii) allows selective exploitation of complementary clinical
signals, and (iii) provides label-specific fusion weights to quantify the contribution of EHR information to each diagnostic endpoint.

% 微调最优超参：lr=1e-5 epoch=3 batch_size=10
% 线性探针超参：lr=5e-5 epoch=4
% tabular-only：lr=5e-5 epoch=3

\section{Implementation Details}
\label{sec:implementation details}

\subsection*{Data organization.} To prevent information leakage, we strictly partition the training, validation, and test sets at the \emph{patient level}, ensuring that MRI scans from the same patient never appear in both splits. Each patient is associated with a unique set of diagnostic labels. For the majority of cases, only a single scan is available per anatomical orientation, whereas a small fraction of patients have multiple scans. Given that multi-scan patients constitute only a minor proportion of the cohort, we treat samples as approximately independent and identically distributed (i.i.d.) when training the orientation-specific multi-label classifier.

When fusing the three anatomical orientations, data organization differs. For patients with exactly one scan per orientation, the three modalities are fused directly.  
Patients with missing orientations are excluded, as such incomplete examinations are rare in clinical practice.  
For patients with more than one scan in any orientation, we enumerate all cross-orientation combinations.  
However, this may create a disproportionately large number of fused samples for some patients; for example, a patient with four scans in each orientation yields \(4^3 = 64\) combinations.

To avoid patient-level imbalance introduced by this enumeration, we apply an inverse-frequency weighting scheme: during training, each fused sample is scaled by the reciprocal of the number of combinations originating from the same patient. During inference, prediction probabilities from all fused samples of a patient are aggregated before producing the final diagnostic output.

For anatomical segmentation, MRI volumes are processed independently within a
single orientation, and no cross-orientation fusion is performed; accordingly,
the combination enumeration and inverse-frequency weighting strategy described
above is not required.

% training 3d unet baseline model using epoch = 10
% training 3d resnet baseline model using epoch = 8
\subsection*{Training baseline models.} We adopt standard volumetric neural networks as baselines for both diagnostic classification
and anatomical segmentation tasks.

For anatomical segmentation, we adopt standard \textbf{3D U-Net} and \textbf{UNETR} architectures implemented using the MONAI framework~\cite{cardoso2022monaiopensourceframeworkdeep}. The 3D U-Net encoder consists of five resolution stages with channel dimensions $(64, 64, 128, 128, 256)$ and strided 3D convolutions for downsampling, with a symmetric decoder based on transposed convolutions and skip connections. UNETR replaces the convolutional encoder with a transformer-based encoder configured with an embedding dimension of 768, 12 self-attention heads, and an MLP dimension of 3072, while retaining a U-Net–style decoder that progressively integrates multi-scale features. Both models were trained from scratch for 10 epochs using a learning rate of $1 \times 10^{-3}$ and a batch size of 10 unless otherwise specified.

For diagnostic classification, we consider two commonly used 3D architectures:
a \textbf{3D U-Net} encoder and a \textbf{3D ResNet-18}.
For the 3D U-Net baseline, bottleneck features (\texttt{mid\_0}) are globally average-pooled
and passed to a linear multi-label classification head.
The U-Net backbone matches the architecture used in the diffusion model, enabling a controlled comparison. For knee, the 3D U-Net baseline is trained for 10 epochs with a learning rate of $1 \times 10^{-4}$, while the 3D ResNet-18 baseline is trained for 8 epochs using the same learning rate.

To evaluate cross-anatomy generalization, both classification baselines are trained under
identical protocols on ankle MRI
(12 epochs for 3D U-Net; 10 epochs for 3D ResNet-18)
and shoulder MRI
(8 epochs for 3D U-Net; 6 epochs for 3D ResNet-18).

All models are trained until the validation performance converges.
The checkpoint corresponding to the stabilized validation performance
is used for final test.
This training and test protocol is applied consistently across
\textbf{all methods and tasks}.

% For each baseline, we select the checkpoint with the best validation performance (Macro-Average AUC-ROC or Dice Similarity Coefficient) and report the corresponding test set results.
% baseline报告的是最优的结果

\subsection*{Pre-training the diffusion model for each MRI orientation.} 

OrthoDiffusion adopts a 3D U-net denoising backbone following a standard discrete-time diffusion formulation.
All diffusion models operate on MRI volumes of size $16 \times 256\times256$ with a single input and output channel. The network uses a base channel width of 64 and a
multiscale channel expansion pattern of $(1,1,2,2,4)$ across resolution levels.
Each resolution stage contains one residual block, and self-attention is enabled
at the $16\times 16$ spatial resolution.
A Gaussian diffusion process with \(T=1000\) timesteps is employed, optimized
with an \(\ell_2\) denoising objective.

% Exponential moving average (EMA) weights are maintained with decay \(0.995\) to improve sampling stability.

All diffusion models are trained for \(2.15\times10^{4}\) optimization steps
with batch size \(B=10\), the
learning rate is scaled linearly as
\(\eta = \eta_0 \times B\) with base \(\eta_0 = 1\times10^{-5}\).

\subsection*{Ablation studies on pooling methods.} To evaluate how spatial aggregation influences the discriminative power of diffusion representations, we conducted an ablation study comparing three pooling strategies: global average pooling (GAP), global local pooling (GLP), and self attention pooling (SAP). All experiments were preformed using features extracted from the bottleneck block ($\texttt{mid\_2}$) at diffusion timestep $t=30$, with pretrained sagittal, coronal, and axial backbones kept \textit{frozen} to isolate the effect of pooling mechanism. A multi-label classifier was trained on the pooled representations using the Adam optimizer with an initial learning rate of $5\times10^{-4}$ and a cosine annealing schedule over 10 epochs. The bottleneck feature maps extracted from the 3D U-Net have spatial dimensions $256 \times 1 \times 16 \times 16$. 
For GLP, the two-layer MLP used to compute local attention weights employs a hidden dimension of $C/8$ (with $C=256$, yielding a hidden size of $32$), while SAP is configured with $h=16$ attention heads.

As summarized in Supplementary Table~\ref{tab:linearprobing-pooling-aucroc} and~\ref{tab:linearprobing-pooling-multilabel}, the choice of pooling strategy has a substantial impact on downstream performance. SAP consistently surpasses GAP and GLP across orientations and most disease categories, yielding higher per-disease AUROC and stronger overall multi-label metrics. The gains highlight SAP’s ability to preserve spatially discriminative patterns within diffusion features.

Based on these results, SAP is adopted as the default aggregation method in all subsequent experiments. Unless otherwise specified, the hyperparameter settings used for SAP under both linear probing and fine-tuning are kept identical to those adopted in the ablation experiments.

\subsection*{Ablation studies on fusion methods.}

We conducted an ablation study to evaluate multiple strategies for integrating
multi-orientation diffusion representations, spanning both feature-level and
label-level fusion paradigms. Specifically, we compared five fusion methods:
\textbf{Simple Concatenation}, \textbf{Linear Addition}, \textbf{Linear Concatenation},
\textbf{Cross-Attention Fusion}, and the proposed \textbf{Multi-plane Adaptive Expert}
(\textbf{MPAE}) module. For all fusion comparisons, diffusion features were extracted from the
$\texttt{mid\_2}$ bottleneck block at a fixed diffusion timestep ($t=30$) using
pretrained and frozen sagittal, coronal, and axial diffusion backbones.
Each orientation-specific feature map was subsequently aggregated using the
SAP module, yielding a $512$-dimensional representation
per orientation.

All fusion strategies operate on these $512$-dimensional pose-specific embeddings
and are summarized as follows:

\begin{enumerate}
  \item \textbf{Simple Concat.}
  Orientation-specific representations are concatenated along the channel
  dimension to form a fused feature
  $
  \mathbf{F}_{\mathrm{fuse}} \in \mathbb{R}^{1536}.
  $

  \item \textbf{Linear Concat / Linear Add.}
  Each orientation-specific embedding is projected into a shared latent space of
  dimension $E=512$ via a learnable linear transformation.
  Linear Concat concatenates the projected embeddings into a $1536$-dimensional
  representation, whereas Linear Add merges them by element-wise summation,
  resulting in a $512$-dimensional fused feature.

  \item \textbf{Cross Attention.}
  The three orientation embeddings are projected into a shared space
  ($E=512$) and fused using cross-attention blocks (4 attention heads),
  allowing each orientation to attend to structural cues from the others.
  The refined features are concatenated to produce a $1536$-dimensional
  representation.

  \item \textbf{MPAE.}
  Independent multi-label classifiers are trained for each orientation, yielding
  \(
  \mathbf{z}\!\in\!\mathbb{R}^{3\times K}
  \) (5-fold CV on train folds). 
  A per-label gating MLP (hidden size \(16\), dropout \(0.1\)) predicts label-specific fusion
  weights, and final logits are obtained by weighted logit aggregation as
  in Eq.~\eqref{eq: MAE}.
\end{enumerate}

For \textbf{Simple Concat}, \textbf{Linear Concat/Add}, and \textbf{Cross Attention}, models were trained using the Adam optimizer with an initial learning rate of $5{\times}10^{-5}$, cosine annealing, and 5 training epochs.  
For \textbf{MPAE}, the orientation-specific classifier were trained with a learning rate of $5{\times}10^{-5}$ for 5 epochs, while the gating network was optimized with
a learning rate of $1\times10^{-3}$, weight decay $1\times10^{-4}$, and 10 epochs.

As summarized in Supplementary Table~\ref{tab:linearprobing-fusion-aucroc} and~\ref{tab:linearprobing-fusion-multilabel}, \textbf{Simple Concat} consistently achieved the best or near-best AUROC across the eight abnormality categories and produced the strongest overall multi-label performance. While MPAE offered superior interpretability and the second-highest accuracy, Simple Concat remained the numerically dominant strategy.

Given its superior and stable performance, Simple Concat is adopted as the default fusion strategy for all subsequent experiments. Unless otherwise specified, the hyperparameter settings used
for Simple Concat under both linear probing and fine-tuning are kept identical to those adopted
in the ablation experiments.

\subsection*{MRI-EHR Multi-modal Fusion.}
The MRI branch uses a linear multi-label classification head. To model structured clinical variables, the EHR branch consists of a lightweight feed-forward projection comprising a fully connected
layer with 64 hidden units, followed by ReLU activation, layer normalization, and dropout (0.2), and a final linear output layer. The same EHR architecture is used in both the \emph{EHR-only} baseline and the multi-modal fusion setting to ensure controlled and comparable evaluation.

All models are optimized using Adam. Both linear probing and fine-tuning are performed for 10 epochs with a learning rate of $1\times10^{-5}$, while the EHR-only baseline is trained for 3 epochs using a learning rate of $5\times10^{-5}$.

\subsection*{OrthoDiffusion pipeline.}

We extract intermediate representations from the \textbf{bottleneck blocks} (\text{mid\_0}, \text{mid\_1}, \text{mid\_2}) of 3D Unet denoising network and systematically evaluate diffusion timesteps of 10, 30, 50, 100, 150, 200, 300, and 500. In contrast to encoder and decoder layers, which operate at higher spatial resolutions and incur substantially  greater memory and computational cost, bottleneck features provide a trade-off between representational richness and efficiency, particularly for 3D multi-view MRI analysis. Therefore, bottleneck blocks are adopted as the default configuration throughout all experiments.

To determine task-optimal diffusion representations, we first identify the top four candidate diffusion timestep-block combinations for each anatomical orientation based on validation performance. All \(4^3\) cross-orientation combinations are then exhaustively enumerated to
construct candidate multi-plane configurations.
For knee abnormality diagnosis under linear probing setting, the best-performing configuration—
\textbf{axial}: \(t=50,\, b=\texttt{mid\_0}\);
\textbf{sagittal}: \(t=50,\, b=\texttt{mid\_0}\);
\textbf{coronal}: \(t=100,\, b=\texttt{mid\_2}\)—
is denoted as \textit{LP-Opt}.
Under the \textbf{fine-tuning} setting, the optimal configuration—
\textbf{axial}: \(t=200,\, b=\texttt{mid\_0}\);
\textbf{sagittal}: \(t=150,\, b=\texttt{mid\_0}\);
\textbf{coronal}: \(t=50,\, b=\texttt{mid\_2}\)—
is denoted as \textit{FT-Opt}. The same procedure is applied to ankle and shoulder MRI, yielding
task- and anatomy-specific optimal configurations for each training setting.

For knee anatomical segmentation tasks, we directly select the best-performing
timestep–block combination for each orientation (Sagittal and Coronal) based on validation performance,
without exhaustive cross-orientation enumeration.
This is because segmentation is performed independently for each orientation
and does not involve multi-plane fusion.

The detailed definitions and selected combinations are summarized in
Supplementary Table~\ref{tab:timestep_block_selection}.

For knee, ankle, and shoulder abnormality diagnosis, linear probing in Stage \textbf{I} is conducted
using the hyperparameter configuration defined above. Fine-tuning follows the
same training protocol, except that a substantially reduced learning rate is
used to preserve the pretrained diffusion representations. Specifically, the
learning rate is set to $5 \times 10^{-4} \times 0.05$, and models are fine-tuned
for five epochs. In Stage \textbf{II}, the training procedure and all hyperparameters remain
identical to those specified above.

% linear probing is performed for
% 20 epochs with a learning rate of $1\times10^{-3}$, while 
For knee anatomical segmentation task, fine-tuning is carried
out for 30 epochs using a learning rate of $1\times10^{-3}$.

\subsection*{Additional experiments.}

For label-efficiency analysis, we subsample the training cohort at the patient level using nested subsets of size 100, 500, 1,000, 2,000, and the full dataset for knee disease classification, and 50, 100, 200, 500, and the full dataset for knee anatomical segmentation, ensuring strict inclusion relationships among subsets. All training hyperparameters and optimization settings follow the configurations described above. OrthoDiffusion is evaluated under the fine-tuning setting (\textit{FT-Opt}) for knee disease classification on the Center A+B+C test dataset, and under \textit{FT-Opt-Sag/Cor} for knee anatomical segmentation on the test dataset.

All experiments are conducted on a single NVIDIA H20-3e data-center GPU
with 143\,GB of HBM memory.

% \begin{table}[htbp]
% \centering
% \caption{Distribution of the number of MRI sequences per patient across training and test sets for each anatomical plane. 
% Counts denote the number of patients with $n$ MRI scans of the same plane.}
% \label{tab:pose_mri_distribution}
% \begin{tabular}{lcccccccc}
% \toprule
% \textbf{Dataset} & \textbf{Plane} & \textbf{1 scan} & \textbf{2 scans} & \textbf{3 scans} & \textbf{4 scans} & \textbf{5 scans} & \textbf{6 scans} & \textbf{9 scans} \\
% \midrule
% \multirow{3}{*}{Train} 
% & Sagittal & 7856 & 434 & 29 & 71 & 9 & 3 & 1 \\
% & Coronal  & 8261 & 359 & 19 & 79 & -- & 2 & -- \\
% & Axial    & 8212 & 380 & 40 & 76 & -- & 4 & -- \\
% \midrule
% \multirow{3}{*}{Test}
% & Sagittal & 1824 & 97 & 4 & 16 & 1 & -- & -- \\
% & Coronal  & 1922 & 74 & 3 & 17 & -- & -- & -- \\
% & Axial    & 1912 & 82 & 7 & 18 & -- & -- & -- \\
% \bottomrule
% \end{tabular}
% \end{table}

\clearpage
\setcounter{table}{0}
\renewcommand{\thetable}{\arabic{table}}
\captionsetup[table]{name=Supplementary Table}
\section*{Supplementary Tables}

\begin{table}[htbp]
\centering
\caption{Summary statistics of knee MRI datasets for sagittal anatomical segmentation.}
\label{tab:sag_knee_seg_stats}
\setlength{\tabcolsep}{6pt}
\begin{tabular}{lccc}
\toprule
\textbf{Characteristic} & \textbf{Training set} & \textbf{Test set} & \textbf{Validation set} \\
\midrule
Total MRI scans & 715 & 119 & 60 \\
Centers & 2 & 2 & 2 \\

\midrule
\textbf{Scanner manufacturer} &  &  &  \\
\quad GE & 432 (60.42\%) & 75 (63.03\%) & 41 (68.33\%) \\
\quad Siemens & -- & 3 (2.52\%) & 1 (1.67\%) \\
\quad Philips & -- & -- & -- \\
\quad United Imaging (UIH) & 283 (39.58\%) & 41 (34.45\%) & 18 (30.00\%) \\

\midrule
\textbf{Magnetic field strength} &  &  &  \\
\quad 3.0 T & 455 (63.64\%) & 81 (68.07\%) & 39 (65.00\%) \\
\quad 1.5 T & 260 (36.36\%) & 38 (31.93\%) & 21 (35.00\%) \\

\midrule
\textbf{Scanning parameters} &  &  &  \\
\quad Repetition time (ms, range) & 1686 (1112--2418) & 1705 (1094--2264) & 1700 (1230--2210) \\
\quad Echo time (ms, range) & 23.3 (16.9--48.0) & 23.6 (16.0--42.0) & 23.7 (19.2--30.6) \\

\quad Slice thickness &  &  &  \\
\quad\quad 3.0 mm & 116 (16.22\%) & 18 (15.13\%) & 10 (16.67\%) \\
\quad\quad 3.5 mm & 482 (67.41\%) & 69 (57.98\%) & 35 (58.33\%) \\
\quad\quad 4.0 mm & 117 (16.36\%) & 32 (26.89\%) & 15 (25.00\%) \\

\midrule
\textbf{Patient characteristics} &  &  &  \\
\quad Age (years, range) & 42.9 (6--76) & 42.3 (11--80) & 40.2 (14--78) \\
\quad Sex (female:male) & 286:429 & 54:65 & 25:35 \\
\quad Weight (kg, range) & 84.7 (30--151) & 82.6 (41--98) & 84.6 (68--90) \\

\bottomrule
\end{tabular}
\end{table}

\begin{table}[htbp]
\centering
\caption{Summary statistics of knee MRI datasets for coronal anatomical segmentation.}
\label{tab:cor_knee_seg_stats}
\setlength{\tabcolsep}{6pt}
\begin{tabular}{lccc}
\toprule
\textbf{Characteristic} & \textbf{Training set} & \textbf{Test set} & \textbf{Validation set} \\
\midrule
Total MRI scans & 802 & 134 & 66 \\
Centers & 2 & 2 & 2 \\

\midrule
\textbf{Scanner manufacturer} &  &  &  \\
\quad GE & 582 (72.57\%) & 85 (63.43\%) & 38 (57.58\%) \\
\quad Siemens & 21 (2.62\%) & 3 (2.24\%) & 5 (7.58\%) \\
\quad Philips & -- & -- & -- \\
\quad United Imaging (UIH) & 199 (24.81\%) & 46 (34.33\%) & 23 (34.85\%) \\

\midrule
\textbf{Magnetic field strength} &  &  &  \\
\quad 3.0 T & 655 (81.67\%) & 81 (60.45\%) & 51 (77.27\%) \\
\quad 1.5 T & 147 (18.33\%) & 53 (39.55\%) & 15 (22.73\%) \\

\midrule
\textbf{Scanning parameters} &  &  &  \\
\quad Repetition time (ms, range) & 1699 (1094--2418) & 1706 (1289--2264) & 1722 (1294--2264) \\
\quad Echo time (ms, range) & 23.2 (16.0--48.0) & 23.1 (17.9--32.0) & 24.1 (19.5--48.0) \\

\quad Slice thickness &  &  &  \\
\quad\quad 3.0 mm & 123 (15.34\%) & 22 (16.42\%) & 11 (16.67\%) \\
\quad\quad 3.5 mm & 532 (66.33\%) & 75 (55.97\%) & 39 (59.09\%) \\
\quad\quad 4.0 mm & 147 (18.33\%) & 37 (27.61\%) & 16 (24.24\%) \\

\midrule
\textbf{Patient characteristics} &  &  &  \\
\quad Age (years, range) & 42.1 (6--80) & 43.4 (11--71) & 41.7 (13--65) \\
\quad Sex (female:male) & 378:424 & 59:75 & 29:37 \\
\quad Weight (kg, range) & 83.8 (30--151) & 84.4 (40--151) & 84.5 (68--100) \\

\bottomrule
\end{tabular}
\end{table}

\begin{sidewaystable}

\centering
\caption{Summary statistics of knee MRI datasets for abnormality classification.}
\label{tab:knee_mri_classification_stats}
\setlength{\tabcolsep}{1pt}
\begin{tabular}{lccccc}
\toprule
\textbf{Characteristic} & 
\textbf{Pre-training set} & 
\textbf{Training set} & 
\textbf{Validation set} & 
\textbf{Test set} & 
\textbf{External test set} \\
\midrule
Total MRI scans & 15,948 & 8,760 & 675 & 1350 & 155 \\
Centers & 3 & 3 & 3 & 3 & 4 \\

\midrule
\textbf{Scanner manufacturer} &  &  &  &  \\
\quad GE & 10,616 (66.57\%) & 5,619 (64.14\%) & 429 (63.56\%) & 869 (64.37\%) & -- \\
\quad Siemens & 1,085 (6.80\%) & 432 (4.93\%) & 32 (4.74\%) & 60 (4.44\%)& 86 (55.48\%) \\
\quad Philips & -- & -- & -- & -- & 27 (17.42\%) \\
\quad United Imaging (UIH) & 4,247 (26.63\%) & 2,709 (30.92\%) & 214 (31.70\%) & 421 (31.19\%)  & 42 (27.09\%) \\

\midrule
\textbf{Patient characteristics} &  &  &  & &  \\
\quad Age (years, range) & 
33.0 (3--87) & 
33.3 (6--85) & 
32.9 (6--87) & 
33.1 (6--87) &
41.2 (12--76) \\

\quad Sex (female:male) & 
5534:10414 & 
3101:5659 & 
235:440 & 
475:875&
79:76 \\

\quad Weight (kg, range) & 
85.1 (15.0--151.0) & 
85.0 (15--165) & 
84.2 (50--110) & 
84.9 (40.8--144)&
82.4 (40.5--165.3) \\
\midrule
\textbf{Magnetic field strength} &  &  &  &  &\\
\quad 3.0 T & 11,348 (71.16\%) & 6,572 (75.02\%) & 516 (76.44\%) & 1,015 (75.19\%)& 113 (72.90\%) \\
\quad 1.5 T & 4,600 (28.84\%) & 2,188 (24.98\%) & 169 (25.04\%) & 335 (24.81\%)& 42 (27.10\%) \\

\midrule
\textbf{Scanning parameters} &  &  &  &  \\
\quad Repetition time (ms, range) & 
2495 (1472--4372) & 
2456 (1472--4080) & 
2453 (1582--3751) & 
2455 (1394--3685) &
2627 (1821--3100) \\

\quad Echo time (ms, range) & 
34.8 (16.5--70.4) & 
34.3 (18.0--70.3) & 
34.0 (23.8--63.8) & 
34.3 (23.8--76.5) &
35.3 (21.2--41.0) \\

\quad Slice thickness &  &  &  &  &\\
\quad\quad 3.0 mm & 1,075 (6.74\%) & 406 (4.63\%) & 33 (4.89\%) & 63 (4.67\%) & 19 (12.26\%) \\
\quad\quad 3.5 mm & 13,861 (86.91\%) & 7,544 (86.12\%) & 590 (87.41\%) & 1,196 (88.59\%) & 99 (63.87\%) \\
\quad\quad 4.0 mm & 1,012 (6.35\%) & 810 (9.25\%) & 52 (7.70\%) & 91 (6.74\%)& 37 (23.87\%) \\

\midrule
\textbf{Disease distribution} &  &  &  \\
\quad Anterior cruciate ligament (ACL) injury & -- & 5000 & 244 & 893 & 93 \\
\quad Posterior cruciate ligament (PCL) injury & -- & 783 & 70 & 116 & 21 \\
\quad Medial meniscus (MM) tear & -- & 4822 & 329 & 795  & 111 \\
\quad Lateral meniscus (LM) tear & -- & 4310 & 309 & 681 & 95 \\
\quad Medial collateral ligament (MCL) injury & -- & 1354 & 68 & 199 & 45 \\
\quad Lateral collateral ligament (LCL) injury & -- & 1161 & 18 & 144 & 53 \\
\quad Patellar dislocation (PD) & -- & 382 & 32 & 56 & 21 \\
\quad Joint effusion (EFFU) & -- & 4010 & 269 & 614 & 137 \\
\bottomrule
\end{tabular}
\end{sidewaystable}

\begin{table}[htbp]
\centering
\caption{Summary statistics of ankle MRI datasets for abnormality classification.}
\label{tab:ankle_mri_stats}
\setlength{\tabcolsep}{1pt}
\begin{tabular}{lccc}
\toprule
\textbf{Characteristic} & \textbf{Training set} & \textbf{Test set} & \textbf{Validation set} \\
\midrule
Total MRI scans & 2178 & 256 & 128 \\
Centers & 3 & 3 & 3 \\

\midrule
\textbf{Scanner manufacturer} &  &  &  \\
\quad GE & 741 (34.02\%) & 85 (33.20\%) & 31 (24.22\%) \\
\quad Siemens & 865 (39.72\%) & 114 (44.53\%) & 94 (73.44\%) \\
\quad Philips & -- & -- & -- \\
\quad United Imaging (UIH) & 572 (26.26\%) & 57 (22.27\%) & 3 (2.34\%) \\

\midrule
\textbf{Patient characteristics} &  &  &  \\
\quad Age (years, range) & 33.2 (6--80) & 35.1 (8--77) & 33.9 (6--72) \\
\quad Sex (female:male) & 1111:1067 & 127:129 & 62:66 \\
\quad Weight (kg, range) & 85.3 (30--120) & 85.0 (40--140) & 86.7 (40--120) \\

\midrule
\textbf{Magnetic field strength} &  &  &  \\
\quad 3.0 T & 2047 (93.99\%) & 239 (93.36\%) & 97 (75.78\%) \\
\quad 1.5 T & 131 (6.01\%) & 17 (6.64\%) & 31 (24.22\%) \\

\midrule
\textbf{Scanning parameters} &  &  &  \\
\quad Repetition time (ms, range) & 2172 (1170--3870) & 2184 (1265--3480) & 2455 (1967--3129) \\
\quad Echo time (ms, range) & 33.8 (10.2--52.3) & 33.7 (21.4--49.3) & 35.0 (28.0--48.5) \\

\quad Slice thickness &  &  &  \\
\quad\quad 3.0 mm & 21 (0.96\%) & 6 (2.34\%) & 2 (1.56\%) \\
\quad\quad 3.5 mm & 2055 (94.35\%) & 221 (86.33\%) & 111 (86.72\%) \\
\quad\quad 4.0 mm & 102 (4.68\%) & 29 (11.33\%) & 15 (11.72\%) \\

\midrule
\textbf{Disease distribution} &  &  &  \\
\quad Anterior talofibular ligament (ATFL) injury & 1851 & 223 & 113 \\
\quad Calcaneofibular ligament (CFL) injury & 1784 & 204 & 111 \\
\quad Achilles tendon rupture (ATR) & 310 & 34 & 8 \\
\quad Osteochondral lesion of the talus (OLT) & 666 & 85 & 49 \\

\bottomrule
\end{tabular}
\end{table}

\begin{table}[htbp]
\centering
\caption{Summary statistics of shoulder MRI datasets for abnormality classification.}
\label{tab:shoulder_mri_stats}
\setlength{\tabcolsep}{1pt}
\begin{tabular}{lccc}
\toprule
\textbf{Characteristic} & \textbf{Training set} & \textbf{Test set} & \textbf{Validation set} \\
\midrule
Total MRI scans & 7167 & 1193 & 597 \\
Centers & 5 & 5 & 5 \\

\midrule
\textbf{Scanner manufacturer} &  &  &  \\
\quad GE & 2028 (28.30\%) & 531 (44.51\%) & 1 (0.17\%) \\
\quad Siemens & 923 (12.88\%) & -- & 256 (42.88\%) \\
\quad Philips & -- & -- & -- \\
\quad United Imaging (UIH) & 4216 (58.83\%) & 662 (55.49\%) & 340 (56.95\%) \\

\midrule
\textbf{Patient characteristics} &  &  &  \\
\quad Age (years, range) & 49.7 (9--93) & 49.3 (11--83) & 49.6 (12--85) \\
\quad Sex (female:male) & 3430:3737 & 562:631 & 296:301 \\
\quad Weight (kg, range) & 83.9 (35--140) & 84.2 (41--130) & 83.2 (55--188) \\

\midrule
\textbf{Magnetic field strength} &  &  &  \\
\quad 3.0 T & 5732 (79.98\%) & 827 (69.32\%) & 579 (96.98\%) \\
\quad 1.5 T & 1435 (20.02\%) & 366 (30.68\%) & 18 (3.02\%) \\

\midrule
\textbf{Scanning parameters} &  &  &  \\
\quad Repetition time (ms, range) & 1518 (646--3442) & 1546.5 (604--3173) & 1410 (1004--3020) \\
\quad Echo time (ms, range) & 28.6 (12.3--66.5) & 29.4 (16.9--56.6) & 26.1 (12.8--63.7) \\

\quad Slice thickness &  &  &  \\
\quad\quad 3.0 mm & 21 (0.29\%) & 3 (0.25\%) & 1 (0.17\%) \\
\quad\quad 3.5 mm & 6698 (93.46\%) & 1066 (89.35\%) & 594 (99.50\%) \\
\quad\quad 4.0 mm & 448 (6.25\%) & 124 (10.39\%) & 2 (0.34\%) \\

\midrule
\textbf{Disease distribution} &  &  &  \\
\quad Supraspinatus tendon (SSP) tear  & 6903 & 1153 & 585 \\
\quad Infraspinatus tendon (ISP) tear & 4143 & 683 & 362 \\
\quad Subscapularis tendon (SSC) tear & 3034 & 463 & 259 \\
\quad Long head of the biceps tendon (LHBT) injury & 1771 & 264 & 155 \\
\quad Adhesive capsulitis (AC) & 2364 & 376 & 204 \\
\quad Long head of the biceps tendon (LHBT) sheath effusion & 713 & 114 & 62 \\
\quad Subacromial--subdeltoid (SASD) bursal effusion  & 5607 & 913 & 459 \\

\bottomrule
\end{tabular}
\end{table}

%%%%
\begin{table}[htbp]
    \centering
    \setlength{\tabcolsep}{3.5pt}
    \caption{Ablation study of pooling strategies for the eight-label \textbf{knee injury prediction} task across three orientations on the Center A+B+C test set. To ensure fair comparison across orientations, all features are extracted from \texttt{block=mid\_2} at \texttt{timestep=30}. 
    Results are reported as per-disease AUROC (\%). Boldface indicates the best result under the same setting in each column.}
    \label{tab:linearprobing-pooling-aucroc}
    \begin{tabular}{llcccccccc}
        \toprule
        \textbf{Orientation} & \textbf{Pooling} & \textbf{ACL} & \textbf{PCL} & \textbf{MM} & \textbf{LM} & \textbf{MCL} & \textbf{LCL} & \textbf{PD} & \textbf{EFFU} \\
        \midrule
        \multirow{3}{*}{Sagittal} 
        & GAP & 68.83 & 56.85 & 60.21 & 61.51 & 75.04 & 70.72 & 82.73 & 78.89 \\
        & GLP & 92.88 & 84.42 & 73.26 & 66.26 & 82.00 & 75.68 & 95.07 & 81.93 \\
        & SAP & \textbf{94.76} & \textbf{91.04} & \textbf{75.99} & \textbf{70.54} & \textbf{82.08} & \textbf{74.89} & \textbf{98.81} & \textbf{84.76} \\
        \midrule
        \multirow{3}{*}{Coronal} 
        & GAP & 74.61 & 58.33 & 65.69 & 61.10 & 78.96 & 73.90 & 84.89 & 81.84 \\
        & GLP & 91.71 & 70.61 & 70.18 & 67.26 & 81.04 & 75.67 & 94.37 & 83.84  \\
        & SAP & \textbf{95.00} & \textbf{83.03} & \textbf{78.94} & \textbf{74.72} & \textbf{87.38} & \textbf{77.53} & \textbf{98.20} & \textbf{83.65}  \\
        \midrule
        \multirow{3}{*}{Axial} 
        & GAP & 71.43 & 57.87 & 64.71 & 59.11 & 76.37 & 72.73 & 84.90 & 75.72 \\
        & GLP & 91.09 & 68.87 & 71.05 & 64.46 & 81.38 & 75.73 & 96.46 & 82.81 \\
        & SAP & \textbf{94.25} & \textbf{86.06} & \textbf{75.83} & \textbf{69.34} & \textbf{88.33} & \textbf{79.34} & \textbf{98.15} & \textbf{84.06} \\
        \bottomrule
    \end{tabular}
\end{table}

\begin{table}[htbp]
    \centering
    \setlength{\tabcolsep}{1pt} 
    \caption{Ablation study of pooling strategies for the eight-label \textbf{knee injury prediction} task across three orientations on the Center A+B+C test set. To ensure fair comparison across orientations, all features are extracted from \texttt{block=mid\_2} at \texttt{timestep=30}. 
    Results are reported as per-disease average precision and overall multi-label metrics (\%). Boldface indicates the best result under the same setting in each column.}
    \label{tab:linearprobing-pooling-multilabel}
    \begin{tabular}{llcccccccc|cccccc}
        \toprule
        \textbf{Orientation} & \textbf{Pooling} & \textbf{ACL} & \textbf{PCL} & \textbf{MM} & \textbf{LM} & \textbf{MCL} & \textbf{LCL} & \textbf{PD} & \textbf{EFFU}
        & \textbf{CP} & \textbf{CR} & \textbf{CF1} & \textbf{OP} & \textbf{OR} & \textbf{OF1} \\
        \midrule
        \multirow{3}{*}{Sagittal} 
        & GAP & 73.10 & 12.90 & 62.95 & 59.36 & 33.40 & 22.71 & 20.03 & 73.49 & 44.63 & 36.98 & 35.32 & 62.56 & 62.57 & 62.57 \\
        & GLP & 94.39 & 50.03 & 76.59 & 64.71 & 44.69 & 29.34 & 61.86 & 77.80 & \textbf{69.96} & 46.76 & 50.99 & 72.08 & 66.51 & 69.19 \\
        & SAP & \textbf{95.94} & \textbf{73.29} & \textbf{79.72} & \textbf{69.11} & \textbf{46.31} & \textbf{28.85} & \textbf{86.25} & \textbf{81.99} & 69.26 & \textbf{62.31} & \textbf{64.91} & \textbf{73.49} & \textbf{72.04} & \textbf{72.76} \\
        \midrule
        \multirow{3}{*}{Coronal} 
        & GAP & 77.32 & 14.06 & 68.54 & 58.66 & 39.35 & 25.35 & 20.47 & 77.22 & 48.39 & 38.05 & 38.37 & 65.59 & 62.21 & 63.86 \\
        & GLP & 93.00 & 22.79 & 73.09 & 66.47 & 44.89 & 24.84 & 49.99 & 79.96 & 68.40 & 43.61 & 46.61 & 71.66 & 65.09 & 68.22 \\
        & SAP & \textbf{96.16} & \textbf{50.06} & \textbf{82.84} & \textbf{73.84} & \textbf{63.94} & \textbf{29.86} & \textbf{77.90} & \textbf{80.08} & \textbf{68.60} & \textbf{60.70} & \textbf{63.81} & \textbf{74.84} & \textbf{72.66} & \textbf{73.74} \\
        \midrule
        \multirow{3}{*}{Axial} 
        & GAP & 74.12 & 12.43 & 68.60 & 58.62 & 32.48 & 21.70 & 20.72 & 70.32 & 39.52 & 34.83 & 32.54 & 61.81 & 60.74 & 61.27 \\
        & GLP & 93.35 & 20.16 & 75.41 & 62.72 & 42.16 & 26.34 & 68.94 & 78.23 & 64.09 & 46.38 & 49.08 & 70.49 & 65.16 & 67.72 \\
        & SAP & \textbf{95.42} & \textbf{54.80} & \textbf{79.00} & \textbf{69.30} & \textbf{61.39} & \textbf{27.04} & \textbf{86.97} & \textbf{81.31} & \textbf{66.97} & \textbf{60.68} & \textbf{63.30} & \textbf{73.27} & \textbf{71.05} & \textbf{72.14} \\
        \bottomrule
    \end{tabular}
\end{table}

%%%

\begin{table}[htbp]

    \centering
    \caption{Comparison of different representation fusion strategies for the eight-label \textbf{knee injury prediction} task on the Center A+B+C test dataset. For timestep–block selection settings, both are \texttt{block=mid\_2} and \texttt{timestep=30}. Results are reported in terms of per-disease AUROC(\%). Boldface indicates the best result in each column.}
    \label{tab:linearprobing-fusion-aucroc}
    \begin{tabular}{lcccccccc}
        \toprule
        \textbf{Methods} & \textbf{ACL} & \textbf{PCL} & \textbf{MM} & \textbf{LM} & \textbf{MCL} & \textbf{LCL} & \textbf{PD} & \textbf{EFFU} \\
        \midrule
        Simple Concat & \textbf{96.62} & \textbf{92.47} & \textbf{81.32} & \textbf{77.24} & \textbf{89.98} & \textbf{82.38} & \textbf{99.49} & 86.08  \\
        Linear Concat & 96.52 & 92.46 & 81.05 & 76.94 & 89.67 & 81.26 & 99.42 & 85.50  \\
        Linear Add & 96.54 & \textbf{92.47} & 80.93 & 76.86 & 89.62 & 81.14 & 99.42 & 85.27   \\
        Cross Attention & 96.54 & 92.22 & 80.63 & 76.38 & 89.43 & 80.11 & 99.19 & 84.50 \\
        
        MPAE & 96.54 & 92.08 & 80.86 & 76.06 & 89.70 & 82.29 & 99.46 & \textbf{86.12}\\
        \bottomrule
    \end{tabular}
\end{table}

\begin{table}[htbp]

    \centering
    \setlength{\tabcolsep}{2pt} % 调小列间距
    \caption{Comparison of different representation fusion strategies for the eight-label \textbf{knee injury prediction} task on the Center A+B+C knee dataset. For timestep–block selection settings, both are \texttt{block=mid\_2} and \texttt{timestep=30}. Results are reported in terms of per-disease AP and overall multi-label metrics (\%). Boldface indicates the best result in each column.}
    \label{tab:linearprobing-fusion-multilabel}
    \begin{tabular}{lcccccccc|cccccc}
        \toprule
        \textbf{Methods} & \textbf{ACL} & \textbf{PCL} & \textbf{MM} & \textbf{LM} & \textbf{MCL} & \textbf{LCL} & \textbf{PD} & \textbf{EFFU} 
        & \textbf{CP} & \textbf{CR} & \textbf{CF1} & \textbf{OP} & \textbf{OR} & \textbf{OF1} \\
        \midrule
        Simple Concat & \textbf{97.32} & \textbf{76.97} & \textbf{85.18} & \textbf{76.79} & \textbf{69.04} & \textbf{36.00} & \textbf{92.30} & 83.40 & 75.24 & 66.39 & \textbf{70.22} & 78.62 & 74.73 & \textbf{76.63}  \\
        Linear Concat & 97.23 & 77.32 & 84.91 & 76.73 & 68.12 & 34.06 & 92.19 & 82.93 & 73.67 & 67.53 & \textbf{70.22} & 77.09 & 75.39 & 76.23  \\
        Linear Add & 97.24 & 76.88 & 84.76 & 76.67 & 67.89 & 33.91 & 91.80 & 82.75 & 73.31 & 67.28 & 69.93 & 76.92 & 75.26 & 76.08  \\
        Cross Attention & 97.29 & 75.77 & 84.45 & 76.12 & 68.05 & 33.08 & 90.47 & 81.76 & 72.48 & \textbf{67.78} & 69.83 & 76.17 & \textbf{75.67} & 75.92 \\
        MPAE & 97.27 & 75.12 & 84.45 & 75.66 & 67.77 & 35.18 & 91.62 & \textbf{83.58} & \textbf{76.28} & 62.85 & 67.89 & \textbf{78.70} & 73.80 & 76.17    \\
        \bottomrule
    \end{tabular}
\end{table}

%%%

\begin{table}[htbp]
\centering
\caption{Task-specific optimal diffusion timestep and network block selections for different musculoskeletal tasks and anatomical sites.}
\label{tab:timestep_block_selection}
\setlength{\tabcolsep}{4pt}
\begin{tabular}{lllcccccc}
\toprule
\multirow{2}{*}{\textbf{Task}} & \multirow{2}{*}{\textbf{Anatomy}} & \multirow{2}{*}{\textbf{Setting}} 
& \multicolumn{2}{c}{\textbf{Sagittal}} 
& \multicolumn{2}{c}{\textbf{Coronal}} 
& \multicolumn{2}{c}{\textbf{Axial}} \\
\cmidrule(lr){4-5} \cmidrule(lr){6-7} \cmidrule(lr){8-9}
 &  &  
 & \textbf{Timestep} & \textbf{Block} 
 & \textbf{Timestep} & \textbf{Block} 
 & \textbf{Timestep} & \textbf{Block} \\
\midrule
\multirow{4}{*}{Classification}
 & Knee      & LP-Opt & 50 & mid\_0 &  100 & mid\_2 & 50 &  mid\_0 \\
 & Knee      & FT-Opt & 150 & mid\_0 & 50  & mid\_2 & 200 &  mid\_0 \\
 & Ankle     & FT-A-Opt & 300  & mid\_0 & 100 & mid\_1 & 30 & mid\_0 \\
 & Shoulder  & FT-S-Opt & 200 & mid\_2 & 30 & mid\_2 & 100 & mid\_1  \\
\midrule
\multirow{1}{*}{Segmentation}
 % & Knee      & LP-Opt-Sag/Cor & 50 & mid\_0 & 30 & mid\_2  & -- & -- \\
 & Knee      & FT-Opt-Sag/Cor & 50 & mid\_2 & 50 & mid\_2 & -- & -- \\
\bottomrule
\end{tabular}
\end{table}

\begin{table}[htbp]
    \centering
    \setlength{\tabcolsep}{2pt}
    \caption{Comparisons of per-disease AUROC (\%) for the eight-label \textbf{knee injury prediction} task on the Center A-C or D-G test dataset. Boldface indicates the best result under the same setting in each column.}
    \label{tab:knee_multilabel_auc_baseline_aucroc}
    \begin{tabular}{l l ccccccccc}
        \toprule
        \textbf{Dataset} & \textbf{Methods} & \textbf{ACL} & \textbf{PCL} & \textbf{MM} & \textbf{LM} & \textbf{MCL} & \textbf{LCL} & \textbf{PD} & \textbf{EFFU} & \textbf{Macro-AUC} \\
        \midrule
        \multirow{4}{*}{Center A+B+C} 
        & 3D-Unet & 96.28 & 88.01 & 72.68 & 69.56 & 84.99 & 79.77 & 95.94 & 86.09 & 84.17 \\
        & 3D-ResNet-18 & 96.96 & 87.56 & 74.90 & 70.81 & 85.00 & 77.82 & 97.49 & 84.41 & 84.37 \\
        & \textbf{Ours (LP, LP-Opt)} & 96.53 & 92.43 & 81.04 & 77.73 & 90.37 & 82.84 & 99.53 & 86.02 & 88.31 \\
        % & \textbf{Ours (Fine-tuning, LP-Opt)} & \textbf{98.16} & \textbf{94.47} & 86.56 & 84.52 & 92.72 & \textbf{85.98} & 99.61 & \textbf{86.43} & 91.18 \\
        & \textbf{Ours (FT, FT-Opt)} & \textbf{98.06} & \textbf{94.35} & \textbf{87.31} & \textbf{85.62} & \textbf{92.97} & \textbf{85.72} & \textbf{99.63} & \textbf{86.39} & \textbf{91.26} \\
        \midrule
        \multirow{4}{*}{Center D+E+F+G}
        & 3D-Unet 
        & 71.21 & 66.88 & 73.35 & 63.49 & 73.60 & 65.38 & 86.06 & 72.92 
        & 71.61 \\
        
        & 3D-ResNet-18 
        & 76.95 & 74.38 & 72.85 & 64.98 & 71.55 & 66.32 & 90.29 & 71.71 
        & 73.63 \\
        
        & \textbf{Ours (LP, LP-Opt)}   
        & 78.10 & \textbf{87.78} & 86.06 & 72.28 & 76.38 & 73.27 & \textbf{95.98} & \textbf{78.95} 
        & 81.10 \\
        
        % & \textbf{Ours (Fine-tune, LP-Opt)} 
        % & 77.04 & 81.63 & 86.08 & 77.12 & 78.16 & 72.38 & 94.74 & 72.06 
        % & 79.65 \\
        
        & \textbf{Ours (FT, FT-Opt)} 
        & \textbf{80.02} & 85.22 & \textbf{89.35} & \textbf{78.65} & \textbf{81.17} & \textbf{73.42} & 95.95 & 76.97 
        & \textbf{82.59} \\
        \bottomrule
    \end{tabular}
\end{table}

\begin{table}[htbp]
    \centering
    \setlength{\tabcolsep}{1pt}
    \caption{Comparisons of per-disease AP and overall multi-label metrics (\%) for the eight-label \textbf{knee injury prediction} task on the Center A+B+C test dataset. Boldface indicates the best result in each column.}
    \label{tab:knee_multilabel_ap_baseline}
    \begin{tabular}{lcccccccc|cccccc}
        \toprule
        \textbf{Methods} & \textbf{ACL} & \textbf{PCL} & \textbf{MM} & \textbf{LM} & \textbf{MCL} & \textbf{LCL} & \textbf{PD} & \textbf{EFFU}
        & \textbf{CP} & \textbf{CR} & \textbf{CF1} & \textbf{OP} & \textbf{OR} & \textbf{OF1} \\
        \midrule
        3D-Unet &  97.14 & 63.96 & 76.55 & 68.80 & 49.67 & 30.22 & 68.18 & 83.15 & 68.52 & 56.39 & 59.98 & 73.54 & 70.88 & 72.18   \\
        3D-Resnet-18 & 97.70 & 65.11 & 79.31 & 68.65 & 54.79 & 29.15 & 79.76 & 81.66 & 68.64 & 62.35 & 63.61 & 71.08 & 76.35 & 73.62  \\
        \textbf{Ours (LP, LP-Opt)} &  97.14 & 78.41 & 85.23 & 77.54 & 69.88 & 37.18 & 92.36 & 82.73 & 75.44 & 67.76 & 71.11 & 78.03 & 75.20 & 76.59   \\
        % \textbf{Ours (Fine-tune, LP-Opt)} & 98.39 & 84.27 & 89.82 & 84.78 & 75.31 & 39.73 & 92.79 & 84.38 & 77.99 & 72.51 & 75.06 & 81.69 & 78.82 & 80.23  \\
        \textbf{Ours (FT, FT-Opt)} & \textbf{98.49} & \textbf{83.38 }& \textbf{90.23} & \textbf{85.93} & \textbf{75.79} & \textbf{40.37} & \textbf{93.74} & \textbf{83.49} & \textbf{79.48}& \textbf{72.92 }& \textbf{75.90} & \textbf{82.36 }& \textbf{79.34} & \textbf{80.82}       \\
        \bottomrule
    \end{tabular}
\end{table}

\begin{table}[htbp]
    \centering
    \setlength{\tabcolsep}{1pt}
    \caption{Comparisons of per-disease AP and overall multi-label metrics (\%) for the eight-label \textbf{knee injury prediction} task on the Center D--G test dataset. Boldface indicates the best result in each column.}
    \label{tab:knee_multicenter_ap}
    \begin{tabular}{lcccccccc|cccccc}
        \toprule
        \textbf{Methods} & \textbf{ACL} & \textbf{PCL} & \textbf{MM} & \textbf{LM} & \textbf{MCL} & \textbf{LCL} & \textbf{PD} & \textbf{EFFU}
        & \textbf{CP} & \textbf{CR} & \textbf{CF1} & \textbf{OP} & \textbf{OR} & \textbf{OF1}  \\
        \midrule
        3D-Unet & 81.15 & 32.21 & 86.10 & 72.63 & 52.68 & 45.46 & 55.89 & 95.70 & 69.35 & 45.33 & 53.62 & 77.85 & 54.96 & 64.43   \\
        3D-Resnet-18 & 83.73 & 33.96 & 86.59 & 71.53 & 50.77 & 45.84 & 71.58 & 95.54 & 70.25 & 46.70 & 53.66 & 79.57 & 57.68 & 66.88 \\
        \textbf{Ours (LP, LP-Opt)}   & 86.06 & 55.13 & 94.00 & 78.25 & 60.59 & 54.69 & \textbf{85.25 }& \textbf{96.70 }& 79.97 & 51.23 & 60.81 & 82.62 & 60.24 & 69.68       \\
        % \textbf{Ours (Fine-tune, LP-Opt)}   & 85.22 & 51.58 & 93.85 & 79.72 & 63.54 & 50.99 & 81.67 & 95.41 & 82.78 & 53.33 & 62.69 & 82.71 & 61.46 & 70.52       \\
        \textbf{Ours (FT, FT-Opt)}   & \textbf{87.60} & \textbf{61.76 }& \textbf{95.36}& \textbf{85.24} & \textbf{68.63} & \textbf{57.98} & 84.79 & 96.40 & \textbf{83.61} & \textbf{56.82} & \textbf{66.18} & \textbf{86.61} & \textbf{65.10} & \textbf{74.33}       \\
        \bottomrule
    \end{tabular}
\end{table}
\begin{table}[htbp]
    \centering
    \setlength{\tabcolsep}{0.8pt}
    \caption{Comparisons of per-disease AP, overall multi-label metrics (\%) on the eight-label \textbf{knee injury prediction} task on the A+B+C+D+E+F+G test dataset with different magnetic field strengths. Boldface indicates the best result under the same setting in each column.}
    \label{tab:knee_multilabel_baseline_ap_field_strengths}
    \begin{tabular}{l l cccccccc|cccccc}
        \toprule
        \textbf{\textbf{\makecell{Field \\ Strength}}} & \textbf{Methods} & \textbf{ACL} & \textbf{PCL} & \textbf{MM} & \textbf{LM} & \textbf{MCL} & \textbf{LCL} & \textbf{PD} & \textbf{EFFU}
        & \textbf{CP} & \textbf{CR} & \textbf{CF1} & \textbf{OP} & \textbf{OR} & \textbf{OF1} \\
        \midrule
        \multirow{3}{*}{1.5T} 
        & 3D-Unet & 95.09 & 55.03 & 70.32 & 66.30 & 58.32 & 44.45 & 63.02 & 86.98 & 68.35 & 55.93 & 59.82 & 73.17 & 64.25 & 68.42 \\
        & 3D-ResNet-18 & 94.55 & 53.57 & 76.06 & 68.19 & 61.24 & 46.35 & 69.50 & 84.43 & 67.59 & 60.24 & 62.83 & 73.38 & 66.14 & 69.57  \\
        & \textbf{Ours (FT, FT-Opt)} & \textbf{97.66} & \textbf{77.42} & \textbf{89.70} & \textbf{83.84} & \textbf{83.09} & \textbf{68.25} & \textbf{89.00} & \textbf{87.03} & \textbf{83.01} & \textbf{71.81} & \textbf{76.63} & \textbf{82.83} & \textbf{76.73} & \textbf{79.67}\\
        \midrule
        \multirow{3}{*}{3T}
        & 3D-Unet 
        & 96.30 & 61.21 & 77.08 & 67.72 & 46.42 & 24.18 & 68.26 & 82.41 & 67.33 & 58.63 & 60.89 & 72.46 & 70.90 & 71.67 \\
        
        & 3D-ResNet-18 
        &  97.08 & 63.17 & 81.07 & 70.41 & 49.33 & 23.17 & 74.87 & 81.35 & 65.95 & 59.92 & 61.98 & 74.50 & 69.01 & 71.65 \\
        
        & \textbf{Ours (FT, FT-Opt)} 
        & \textbf{98.14} & \textbf{83.63} & \textbf{90.62} & \textbf{86.29} & \textbf{71.19} & \textbf{35.60} & \textbf{93.29} & \textbf{84.18} & \textbf{79.12} & \textbf{70.33} & \textbf{74.28} & \textbf{82.77} & \textbf{77.97} & \textbf{80.29}  \\
        \bottomrule
    \end{tabular}
\end{table}

\begin{table}[htbp]
    \centering
    \setlength{\tabcolsep}{2pt}
    \caption{Comparisons of per-disease AP, overall multi-label metrics (\%) for the four-label \textbf{ankle injury prediction} task on the test dataset. Boldface indicates the best result in each column.}
    \label{tab:ankle_multicenter_ap}
    \begin{tabular}{lcccc|cccccc}
        \toprule
        \textbf{Methods} & \textbf{ATFL} & \textbf{CFL} & \textbf{ATR} & \textbf{OLT} 
        & \textbf{CP} & \textbf{CR} & \textbf{CF1} & \textbf{OP} & \textbf{OR} & \textbf{OF1}  \\
        \midrule
        3D-Unet & 91.94 & 89.68 & 32.61 & 42.37 & 67.02 & 49.94 & 45.77 & \textbf{84.06} & 78.01 & 80.92  \\
        3D-Resnet-18 & 93.04 & 90.00 & 29.45 & 54.55 & 61.45 & \textbf{69.77} & \textbf{63.55} & 73.93 & \textbf{87.04} & 79.95  \\
        \textbf{Ours (FT, FT-A-Opt)}   & \textbf{95.73} & \textbf{93.49} & \textbf{40.90} & \textbf{72.26} & \textbf{85.46} & 60.82 & 60.59 & 83.36 & 83.52 & \textbf{83.44}     \\
        \bottomrule
    \end{tabular}
\end{table}

% \begin{table}[htbp]
%     \centering
%     \setlength{\tabcolsep}{0.5pt}
%     \caption{\textbf{Shoulder test} results on a multi-center dataset. 
%     Comparisons of per-disease AP, overall multi-label metrics (\%) are reported for the eight shoulder injury prediction tasks.}
%     \label{tab:shoulder_multicenter_ap}
%     \begin{tabular}{lcccccccc|cccccc}
%         \toprule
%         \textbf{Methods} & \textbf{SSP} & \textbf{ISP} & \textbf{SSC} & \textbf{BT} 
%         & \textbf{AC} 
%         & \textbf{GHJ} & \textbf{LHBT} & \textbf{SASD} 
%         & \textbf{CP} & \textbf{CR} 
%         & \textbf{CF1} & \textbf{OP} 
%         & \textbf{OR}
%         & \textbf{OF1}  \\
%         \midrule
%         3D-Unet & 99.19 & 80.19 & 65.05 & 54.97 & 74.47 & 76.26 & 17.17 & 92.87 & 65.58 & 60.19 & 61.37 & 79.73 & 75.86 & 77.75  \\
%         3D-Resnet-18 & 99.32 & 80.32 & 66.85 & 55.59 & 75.53 & 77.21 & 18.27 & 92.59 & 64.37 & 63.69 & 63.08 & 77.73 & \textbf{79.69 }& 78.70  \\
%         \textbf{Ours (FT, FT-S-Opt)}   &  \textbf{99.54 }& \textbf{84.10 }& \textbf{74.27 }& \textbf{63.20 }& \textbf{82.76 }& \textbf{78.67} & \textbf{30.73} & \textbf{94.06} & \textbf{72.94 }& \textbf{66.66 }& \textbf{67.86 }& \textbf{81.00 }& 79.57 & \textbf{80.28}    \\
%         \bottomrule
%     \end{tabular}
% \end{table}
\begin{table}[htbp]
    \centering
    \setlength{\tabcolsep}{0.5pt}
    \caption{Comparisons of per-disease AP, overall multi-label metrics (\%) for the seven-label \textbf{shoulder injury prediction} task on the test dataset. Boldface indicates the best result in each column.}
    \label{tab:shoulder_multicenter_ap}
    \begin{tabular}{lccccccc|cccccc}
        \toprule
        \textbf{Methods} & \textbf{SSP} & \textbf{ISP} & \textbf{SSC} & \textbf{\makecell{LHBT \\ injury}} 
        & \textbf{AC}  & \textbf{\makecell{LHBT \\ effusion}} & \textbf{SASD} 
        & \textbf{CP} & \textbf{CR} 
        & \textbf{CF1} & \textbf{OP} 
        & \textbf{OR}
        & \textbf{OF1}  \\
        \midrule
        3D-Unet & 99.19 & 80.19 & 65.05 & 54.97 & 74.47 & 17.17 & 92.87 & 65.58 & 60.19 & 61.37 & 79.73 & 75.86 & 77.75  \\
        3D-Resnet-18 & 99.32 & 80.32 & 66.85 & 55.59 & 75.53 & 18.27 & 92.59 & 64.37 & 63.69 & 63.08 & 77.73 & \textbf{79.69 }& 78.70  \\
        \textbf{Ours (FT, FT-S-Opt)}   &  \textbf{99.54 }& \textbf{84.10 }& \textbf{74.27 }& \textbf{63.20 }& \textbf{82.76 } & \textbf{30.73} & \textbf{94.06} & \textbf{72.94 }& \textbf{66.66 }& \textbf{67.86 }& \textbf{81.00 }& 79.57 & \textbf{80.28}    \\
        \bottomrule
    \end{tabular}
\end{table}

\begin{table}[htbp]
    \centering
    \setlength{\tabcolsep}{2pt}
    \caption{Comparisons of per-disease AUROC(\%) for the four-label \textbf{ankle injury prediction} task on the test dataset. Boldface indicates the best result in each column.}
    \label{tab:ankle_multicenter_auc}
    \begin{tabular}{lcccc}
        \toprule
        \textbf{Methods} & \textbf{ATFL} & \textbf{CFL} & \textbf{ATR} & \textbf{OLT} \\
        \midrule
        3D-Unet & 65.31 & 68.93 & 74.88 & 61.06 \\
        3D-Resnet-18 & 66.62 & 69.16 & 75.78 & 68.54  \\
        \textbf{Ours (FT, FT-A-Opt)}   &  \textbf{77.52} & \textbf{77.86} & \textbf{82.00} & \textbf{80.43}    \\
        \bottomrule
    \end{tabular}
\end{table}

% \begin{table}[htbp]
%     \centering
%     \setlength{\tabcolsep}{2pt}
%     \caption{\textbf{Shoulder test} results on a multi-center dataset. 
%     Comparisons of per-disease AUROC(\%) are reported for the eight shoulder injury prediction tasks.}
%     \label{tab:shoulder_multicenter_auc}
%     \begin{tabular}{lcccccccc}
%         \toprule
%         \textbf{Methods} & \textbf{SSP} & \textbf{ISP} & \textbf{SSC} & \textbf{BT} 
%         & \textbf{AC} 
%         & \textbf{GHJ} & \textbf{LHBT} & \textbf{SASD} \\
%         \midrule
%         3D-Unet & 81.72 & 74.59 & 73.66 & 78.25 & 86.00 & 68.85 & 65.82 & 80.25   \\
%         3D-Resnet-18 & 83.48 & 74.92 & 74.97 & 78.40 & 86.30 & 69.19 & 67.03 & 79.01  \\
%         \textbf{Ours (FT, FT-S-Opt)}   & \textbf{88.49} & \textbf{79.08} & \textbf{79.54} & \textbf{81.49} & \textbf{91.03} & \textbf{70.52} & \textbf{75.77} & \textbf{83.38}      \\
%         \bottomrule
%     \end{tabular}
% \end{table}

\begin{table}[htbp]
    \centering
    \setlength{\tabcolsep}{2pt}
    \caption{Comparisons of per-disease AUROC(\%) for the seven-label \textbf{shoulder injury prediction} task on the test dataset. Boldface indicates the best result in each column.}
    \label{tab:shoulder_multicenter_auc}
    \begin{tabular}{lccccccc}
        \toprule
        \textbf{Methods} & \textbf{SSP} & \textbf{ISP} & \textbf{SSC} & \textbf{\makecell{LHBT \\ injury}} 
        & \textbf{AC}  & \textbf{\makecell{LHBT \\ effusion}}  & \textbf{SASD} \\
        \midrule
        3D-Unet & 81.72 & 74.59 & 73.66 & 78.25 & 86.00  & 65.82 & 80.25   \\
        3D-Resnet-18 & 83.48 & 74.92 & 74.97 & 78.40 & 86.30 & 67.03 & 79.01  \\
        \textbf{Ours (FT, FT-S-Opt)}   & \textbf{88.49} & \textbf{79.08} & \textbf{79.54} & \textbf{81.49} & \textbf{91.03} & \textbf{75.77} & \textbf{83.38}      \\
        \bottomrule
    \end{tabular}
\end{table}
\begin{sidewaystable}
    \centering
    \setlength{\tabcolsep}{2.5pt}
    \caption{Comparison of per-disease AP, overall multi-label metrics (\%) for the eight-label \textbf{knee injury prediction} task using different modalities (MRI-only, EHR-only, and MRI+EHR) on the Center A+B+C+D+E+F+G test set. 
    Boldface indicates the best result under same setting in each column.}
    \label{tab:knee_multilabel_ap_merged}
    \begin{tabular}{l l l cccccccc|cccccc}
        \toprule
        \textbf{Dataset} & \textbf{Modality} & \textbf{Methods} 
        & \textbf{ACL} & \textbf{PCL} & \textbf{MM} & \textbf{LM} & \textbf{MCL} & \textbf{LCL} & \textbf{PD} & \textbf{EFFU}
        & \textbf{CP} & \textbf{CR} & \textbf{CF1} & \textbf{OP} & \textbf{OR} & \textbf{OF1}\\
        \midrule

        \multirow{9}{*}{\shortstack{Center A+B+\\C+D+E+F+G}} 

        % -------- MRI only --------
        & \multirow{4}{*}{MRI-only}
        & 3D-Unet & 95.92 & 59.30 & 75.74 & 67.25 & 49.17 & 30.24 & 65.99 & 83.44 & 67.42 & 57.75 & 60.49 & 72.59 & 68.93 & 70.71 \\
        & & 3D-ResNet-18 & 96.14 & 55.90 & 79.40 & 69.71 & 52.62 & 28.70 & 70.20 & 81.62 & 63.50 & 63.08 & 62.80 & 70.55 & 73.93 & 72.20 \\
        & & \textbf{Ours (LP, LP-Opt)} & 96.31 & 74.47 & 85.53 & 76.47 & 67.68 & 34.80 & 89.01 & 84.76 & 75.54 & 64.27 & 68.98 & 78.52 & 73.05 & 75.68 \\
        & & \textbf{Ours (FT, FT-Opt)} & \textbf{97.92} & \textbf{81.56} & \textbf{90.44} & \textbf{85.73} & \textbf{74.64} & \textbf{44.68} & \textbf{91.39} & \textbf{84.96} & \textbf{79.52} & \textbf{70.85} & \textbf{74.69} & \textbf{82.18} & \textbf{77.59} & \textbf{79.82} \\

        \cmidrule(l){2-17}

        % -------- EHR only --------
        & EHR-only
        & EHR-only
        & 58.21 & 11.09 & 56.58 & 49.63 & 16.75 & 19.53 & 7.04 & 56.43 & 33.09 & 32.53 & 29.08 & 55.64 & 55.71 & 55.67 \\

        \cmidrule(l){2-17}

        % -------- Multi-modal --------
        & \multirow{2}{*}{EHR + MRI}
        & \textbf{Ours (LP, LP-Opt)}
        &  96.39 & 73.61 & 85.31 & 76.34 & 67.49 & 37.96 & 88.41 & 85.23 & 78.31 & 62.42 & 68.54 & 79.67 & 72.47 & 75.90\\

        & & \textbf{Ours (FT, FT-Opt)}
        & \textbf{97.89} & \textbf{81.79} & \textbf{90.42} & \textbf{85.70} & \textbf{74.60} & \textbf{46.06} & \textbf{91.42} & \textbf{85.50} & \textbf{81.17} & \textbf{69.23} & \textbf{74.40} & \textbf{83.28} & \textbf{76.81} & \textbf{79.91}  \\

        \bottomrule
    \end{tabular}
\end{sidewaystable}

\end{appendices}

%%===========================================================================================%%
%% If you are submitting to one of the Nature Portfolio journals, using the eJP submission   %%
%% system, please include the references within the manuscript file itself. You may do this  %%
%% by copying the reference list from your .bbl file, paste it into the main manuscript .tex %%
%% file, and delete the associated \verb+\bibliography+ commands.                            %%
%%===========================================================================================%%
\newpage
\clearpage
\bibliography{sn-bibliography}% common bib file
%% if required, the content of .bbl file can be included here once bbl is generated
%%\input sn-article.bbl

\end{document}